\documentclass{article}

\usepackage{arxiv}

\usepackage[utf8]{inputenc} 
\usepackage[T1]{fontenc}    
\usepackage{hyperref}       
\usepackage{url}            
\usepackage{booktabs}       
\usepackage{amsfonts}       
\usepackage{nicefrac}       
\usepackage{microtype}      
\usepackage{lipsum}

\usepackage{subfiles}
\usepackage{tikz}
\usepackage{mathrsfs}
\usepackage{amsmath}
\usepackage{pgfplots}
\usepackage{adjustbox}
\usepackage{multirow}
\usepackage{sansmath}
\usepackage{dsfont}
\usepackage{rotating}
\usepackage{subfig}
\usepackage{longtable}
\usepackage{url}
\usepackage{appendix}
\usepackage{algpseudocode, algorithm}

\definecolor{blue_1}{RGB}{73, 165, 230}
\definecolor{red_1}{RGB}{185, 42, 69}

\title{Copying Machine Learning Classifiers}

\author{
  Irene Unceta\\
  BBVA Data \& Analytics\\
  Universitat de Barcelona \\
  Barcelona, Spain \\
  \texttt{irene.unceta@bbvadata.com} \\
   \And
  Jordi Nin \\
  Dept. of Operations, Innovation and Data Science\\
  ESADE Universitat Ramon Llull\\
  Barcelona, Spain \\
  \texttt{jordi.nin@esade.edu} \\
  \And
  Oriol Pujol \\
  Dept. of Mathematics and Computer Science \\
  Universitat de Barcelona \\
  Barcelona, Spain \\
  \texttt{oriol\_pujol@ub.edu} \\
}

\begin{document}

\twocolumn[
  \begin{@twocolumnfalse}
    
    \maketitle
    
    \begin{abstract}
    We study model-agnostic copies of machine learning classifiers, new models that replicate the decision behavior of any classifier. We develop the theory behind the problem of copying, highlighting its differences with that of learning, and propose a framework to copy the functionality of any classifier using no prior knowledge of its parameters or training data distribution. We validate this framework through extensive experiments using data from a series of well-known problems. To further validate this concept, we use three different use cases where desiderata such as interpretability, fairness or productivization constrains need to be addressed. Results show that copies can be exploited to enhance existing solutions and improve them adding new features and characteristics. 
    \end{abstract}
    
    
    \vspace{1cm}
    
    \end{@twocolumnfalse}
]

\section{Introduction}
\label{sec:introduction}

In many every-day examples, performance of state-of-the-art machine learning is held back by operational constraints. Either the data or the models themselves are subject to privacy restrictions \cite{Fredrikson2015ModelCountermeasures}, \cite{Shokri2017MembershipModels}, \cite{Song2017MachineMuch} or specific regulations apply that require models to be self-explanatory \cite{EuropeanParliament2016REGULATIONRegulation}, \cite{Goodman2017EuropeanExplanation}, \cite{Selbst2017MeaningfulExplanation} or fair with respect to sensitive data attributes \cite{Barocas2016BigImpact}, \cite{Friedman1996BiasSystems}, \cite{Hardt2014HowUnfair}. 
Other issues include time or space limitations for deployment, and production bottlenecks in delivering certain models to the market \cite{Sculley2015HiddenSystems}. To the best of our knowledge, these issues have been traditionally addressed by means of tailored solutions. As a result, off-the-shelf machine learning techniques often yield only sub-optimal results.

Under such circumstances, training a new model may seem straightforward. However, a re-training is not always possible, nor advisable. This may be, for example, because production protocols require the maintenance of predictive performance over time, because the specifics of the model are unknown or even because the training data are no longer available. Whatever the cause, the impossibility of re-training calls for new ways to address this situation.

In this article we study copying, the problem of building a new model that replicates the decision behavior of another. The idea of approximating a model's decision boundary can be found in the literature under different topics, including distillation \cite{Hinton2015Distilling,Caruana2006}, model extraction \cite{Craven1995ExtractingNetworks, Tramer2016StealingAPIs} or adversarial learning \cite{Papernot2017PracticalLearning,Lowd2005AdversarialLearning}. In all cases, this notion is introduced on a simplified case-by-case basis, devoid of theoretical foundation. In contrast, we approach this problem from a higher level of abstraction and mathematically frame it under the copying theory. 

For this purpose, we envisage the most general scenario, where we make the minimum number of required assumptions about the amount of information available during the process. In particular, we assume access to the model is limited to a membership query interface. Unlike previous articles, where the training data distribution is directly \cite{Hinton2015Distilling} or indirectly~\cite{Caruana2006} known and where rich information outputs can be used as soft targets for the new model \cite{Liu2018Teacher,Yang2018KnowledgeDI}, we also assume the training data to be lost and the query interface to produce only hard predictions. 

In this context, we propose copying as a methodology to project the decision function learned by a model onto a new hypothesis space that enables the same decision behaviour, while incorporating new features and properties. This process is one of \textit{differential replication}\cite{Darwin}. Copies not only retain the original accuracy, but can also be used to endow classifiers with new characteristics, such as interpretability, online learning or equity features, which may prove useful to overcome the aforementioned limitations.

We summarize the main contributions of this paper as: 

\begin{itemize}

    \item We formalize the problem of building a copy that replicates the decision behavior of a machine learning model in the most general setting.
    
    \item We explore the theoretical implications of copying and show that this problem differs from that of traditional machine learning.

    \item We put this theory into practice to highlight the specific characteristics of copying and validate this proposal on a series of well known problems. 
    
    \item We further illustrate the value of copying for differential replication in three real use cases. First, we address the issues of non-decomposability and delayed time-to-market delivery in non-client mortgage risk scoring. Second, we build an online copy that recovers a critical operating point in a loan default prediction problem. Finally, we use copies to ensure a fair classification of superhero alignment. 
\end{itemize}

The rest of this article is organized as follows. Sec.~\ref{sec:related} presents a literature survey of related work. The theoretical basis for copying is introduced in Sec.~\ref{sec:framework}, while Sec.~\ref{sec:insights} extracts meaningful insights for a practical implementation. In Sec.~\ref{sec:experiments} we validate copies on various UCI problems. In Sec.~\ref{sec:discussion} we consider the advantages and limitations of this methodology and present three real applications. The paper concludes with a brief summary of our findings and an outline of future research.

\section{Related work}
\label{sec:related}

The idea of copying is not new in the literature. We find this notion in early works on concept extraction, where trained artificial neural networks are compiled into a set of representative rules \cite{Andrews1995SurveyANNs}, \cite{Craven1993LearningNetworks}, \cite{Fu1991RuleNets.}, \cite{Thrun1995ExtractingRepresentations}. More recently, distillation has been proposed to transfer the knowledge acquired by a large, complex model (teacher) to a faster, simpler architecture (student) \cite{Hinton2015Distilling, Caruana2006}. 
Papers in this field have explored different forms of supervision from the teacher \cite{Szegedy2016Rethinking}, training the same network in generations \cite{Funarlanello2018Born} or inducing teacher signals with a softened label distribution to convey useful task-dependent information to students \cite{Yang2018KnowledgeDI}. These can all be understood as a form of data enhancement, where rich information outputs by the complex model are used as soft targets to improve the predictive performance of the student. All these articles use similar concepts to that of copying. The aim of copying is not to enable a simple model to learn a complex task, but to ensure the exact replication of a decision boundary.

An important degree of freedom in distillation is the transfer set used to train the simpler model. Traditionally, knowledge transfer has been treated as a standard learning process, where the training data are relabelled and extended to learn an alternative model \cite{Bologna2018ASVMs}. In most cases, the same set is used to train teacher and student, either in its raw form \cite{Bologna2018ASVMs},\cite{Hinton2015Distilling},\cite{Che2016InterpretablePrediction} or enriched with additional synthetic data \cite{Bastani2018Inter},\cite{Craven1995ExtractingNetworks},\cite{Liu2018Teacher}. Some works advocate the use of unlabelled data \cite{Caruana2006,Zeng2000}, extracted from the estimated density of the attributes. In other examples, teachers and students faced with the same task have different access to training data~\cite{Hong2019RDPD}. In this paper, the training data are assumed to be lost and their distribution unknown. What is more, model internals remain secret throughout the process. 

A seemingly related but vastly different approach is that of \textit{transferability}-based adversarial learning \cite{Papernot2017PracticalLearning},\cite{Papernot2016TransferabilitySamples},\cite{Liu2017DelvingAttacks},\cite{Szegedy2014IntriguingNetworks}\cite{Biggio2013Evasion}, where a malicious adversary exploits samples crafted from a local substitute of a model to compromise it. In this context, copying does bear a similarity to gray-box attacks in settings involving surrogate learners with limited-knowledge~\cite{Biggio2015}. Note, however, that copies are aimed at replicating the original classification boundary globally. Moreover, the objective of adversarial learning fundamentally differs from ours. An adversary benefits from acquiring knowledge about a model to fool it. Copies are global models that replicate a learned decision behavior.

Copying may use a synthetic sample generator process. This process
shares some similarities with active learning. In general, the objective of active learning is to learn a target function using the minimum number of queries in situations where there is a high cost associated to querying/labelling, as is the case of human annotation~\cite{settles.tr09}. In contrast, when generating synthetic samples for copying the cost of querying a model is negligible. Query minimization in this context could still be desirable. Yet, it is not necessary. In addition, while most query optimization strategies rely on class probability outputs~\cite{Fujii1998Selective}, \cite{Lewis1994},\cite{Lindenbaum2004Selective},\cite{Scheffer2001}, this information is not available during the more general copying scenario.

All in all, the above could be understood as narrow examples of copying in restricted scenarios and with very specific objectives. However, to date, this technique has lacked a more general formal framework. To our knowledge the only work that studies distillation from a theoretical perspective is \cite{LopezPaz2016} and, more recently, \cite{Phuong19}. Yet, both focus on learning using privileged information, as opposed to the label-based approach proposed in this article. Hence, our contribution on top of this body of work is to formalize copying as a problem that differs from that of learning and to highlight its general features and characteristics.

At this point, given the many links with related topics, we may ask ourselves questions such as why do we not  at improving the performance of the model? or why is an exact replica important? Or even, we can wonder why is this work focused on a data-less black-box scenario when we usually have training data?

\section{Copying}
\label{sec:framework}

\tikzset{
    model/.style={
           circle,
           },
    legend/.style={
            rectangle,
            text centered,
            },
}

\begin{figure}[!t]
\centering
\begin{tikzpicture}[scale=0.50]
    \draw[red_1, very thick, fill=red_1, fill opacity=0.3]  plot[smooth, tension=.7] coordinates {(-3.5,0.5) (-3,2.5) (-1,3.5) (1.5,3) (4,3.5) (4.8,2.5) (4.5,0.5) (2.5,-2) (0,-1.5) (-3,-1) (-3.5,0.5)};
    
\node[draw=blue_1, circle, inner sep=0pt, minimum size=0.2cm, radius=0.2cm, fill=blue_1] at (-6, 3) (ORIGINAL) {};

\node[legend, left of=ORIGINAL, xshift=0.4cm] (OR) {\textbf{\textcolor{blue_1}{$f_\mathcal{O}$}}};

\node[draw=red_1, circle, inner sep=0pt, minimum size=0.2cm, radius=0.2cm, fill=red_1] at (-3.0, 2.5) (OPTIMAL) {};

\node[legend, right of=OPTIMAL, xshift=-0.5cm] (OP) {\textbf{\textcolor{red_1}{$f_\mathcal{C}^*$}}};

\node[draw=red_1, circle, inner sep=0pt, minimum size=0.2cm, radius=0.2cm, fill=red_1] at (1.6, -0.7) (COPY) {};

\node[legend, right of=COPY, xshift=-0.5cm] (cop) {\textbf{\textcolor{red_1}{$f_{\mathcal{C}}$}}};

\node[legend, right of=COPY, xshift=0.5cm, yshift=-0.5cm] (spa) {\textbf{\textcolor{red_1}{$\mathcal{H}_{\mathcal{C}}$}}};

\path (ORIGINAL) edge[black, dashed, line width=0.3mm] node[above, pos=0.9, outer sep=6pt]{} (OPTIMAL)
(ORIGINAL) edge[black, dashed, line width=0.3mm] node[below,pos=0.4,outer sep=3pt]{} (COPY); 

\end{tikzpicture}
\caption{Copying as a projection of a decision function onto a new hypothesis space $\mathcal{H}_C$.  This space need not coincide with that of the  classifier, i.e. $f_\mathcal{O}$ and $f_\mathcal{C}$ need not belong to the same family of models, and they most usually don't. The optimal copy $f_{\mathcal{C}}^*$ is the closest to $f_\mathcal{O}$.}\label{fig:hypothesis}
\end{figure}
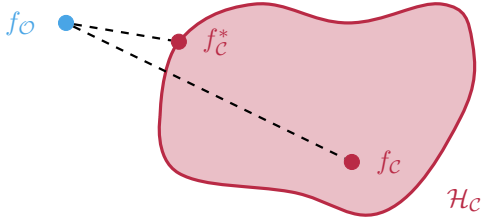

Copying refers to the process of building a functional model which is equivalent in its decision behaviour to another. During this process, the knowledge acquired by the first model is transferred to a copy, in circumstances where both the internals and the training data of the former are unknown, and access to its knowledge is only possible through a membership query interface. 

Let us take a classifier $f_\mathcal{O}: \mathcal{X} \to \mathcal{T}$, where $\mathcal{X}$ and $\mathcal{T}$ correspond to the input and label spaces, respectively. We define the set $\mathscr{D} = \{ (\boldsymbol{x}_i, t_i) \}_{i=1}^M$ as the training data, for $M$ the total number of instances, and restrict to the case of classification, where $\mathcal{T} \in \mathbb{Z}_k$ for $k$ the number of classes. 

Copying is defined as the problem of finding a model $f_\mathcal{C}(\theta) \in \mathcal{H}_C$, parameterized by $\theta$, such that given a new sample $\boldsymbol{x}^*$ it predicts the output $y^* = f_{\mathcal{O}}(\boldsymbol{x}^*)$. Our objective is therefore to obtain a new model, the copy, whose decision function mimics that of $f_\mathcal{O}$ all over the space. 

The process of copying can be interpreted as projecting the decision function $f_\mathcal{O}$ onto the new hypothesis space $\mathcal{H}_C$ the copy belongs to. A graphical illustration of this is shown in Fig.~\ref{fig:hypothesis}. As we will later explain in more detail, this new hypothesis space need not coincide with that of $f_\mathcal{O}$. On the contrary, we can exploit to our advantage the fact that both spaces are different to endow the model with new features, not present in the original hypothesis space. This {\it differential replication} process is the crucial characteristic of copying. 

The problem of copying is characterized by the predictive distribution $P(y^*|f_{\mathcal{O}},\boldsymbol{x}^*)$. Marginalizing with respect to the copy parameters $\theta$ 

\begin{displaymath}
P(y^*|f_{\mathcal{O}},\boldsymbol{x}^*) = \int_{\theta \in \mathcal{H}_C} P(t^*| \theta,f_{\mathcal{O}},\boldsymbol{x}^*)P(\theta|f_{\mathcal{O}},\boldsymbol{x}^*)d\theta,
\end{displaymath}

\noindent
for $\mathcal{H}_C$ the complete parameter space for the copy. We simplify this expression by making two basic assumptions. 

First, when building the copy, knowledge about the unseen data point $\boldsymbol{x}^*$ is not available, so that $P(\theta|f_{\mathcal{O}},\boldsymbol{x}^*) = P(\theta|f_{\mathcal{O}})$. Second, once having built the copy, \textit{i.e.} fixed the value of $\theta$, interaction with the classifier $f_\mathcal{O}$ is no longer required, so that $P(y^*|\theta, f_{\mathcal{O}},\boldsymbol{x}^*) = P(y^*|\theta,\boldsymbol{x}^*)$. On this basis, we rewrite the expression above as

\begin{displaymath}
P(y^*|f_{\mathcal{O}},\boldsymbol{x}^*) = \int_{\theta \in \mathcal{H}_C} P(y^*| \theta,\boldsymbol{x}^*)P(\theta|f_{\mathcal{O}})d\theta.    
\end{displaymath}

We take a \textit{winner takes it all} approach and force the posterior to have the form of a point mass density, $P(\theta|f_{\mathcal{O}}) = \delta(\theta - \theta^*)$, for $\delta(.)$ the Dirac delta function and $\theta^*$ the optimal parameter set. All the probability mass is then placed onto $\theta^*$, so that

\begin{displaymath}
P(y^*|f_{\mathcal{O}},\boldsymbol{x}^*) = P(y^*| \theta^*,\boldsymbol{x}^*).   
\end{displaymath}

Hence, the problem of copying can be understood as that of finding the optimal parameter values $\theta^*$ to maximize the posterior probability
\begin{equation}
    \theta^* = \arg\max_{\theta} P(\theta|f_{\mathcal{O}}).
\label{eq:copy_inference}
\end{equation}

\subsection{The need for unlabelled data}

We study the most general scenario, where the training data $\mathscr{D}$ is assumed to be lost. Solving (\ref{eq:copy_inference}) therefore requires that we generate new data in order to gain information about the form of $f_{\mathcal{O}}$ throughout the input space $\mathcal{X}$. We introduce unlabelled data points $\boldsymbol{z} \in \mathcal{X}$ and rewrite (\ref{eq:copy_inference}) as

\begin{equation}
    \theta^* = \arg\max_{\theta} \int_{\boldsymbol{z}\sim P_Z} P(\theta|f_{\mathcal{O}}(\boldsymbol{z})) dP_Z,
\label{eq:copy_inference_log_int}
\end{equation}
\noindent
for an arbitrary generating probability distribution $P_Z$ from which the new samples are independently drawn. This distribution defines the spatial support for the copy, \textit{i.e.} its plausible operational space. In the existing literature, the training data distribution, $P$, is directly \cite{Hinton2015Distilling} or indirectly \cite{Caruana2006} accessible. Here we completely lack this information, so that we cannot match $P_Z$ to our estimate of $P$. Nonetheless, note that despite $P_Z$ could be related to the training distribution, this is not mandatory for our purposes. 

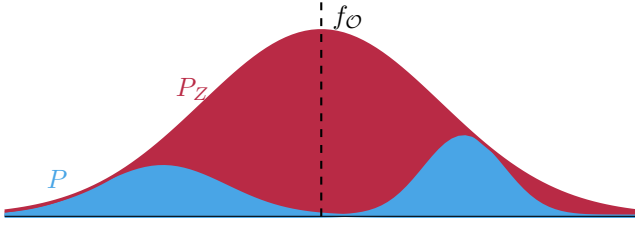
\begin{figure}[!t]
\centering
\begin{tikzpicture}[]
\begin{axis}[
  no markers, domain=0:8, samples=100,
  axis lines*=left, 
  axis y line=none,
  every axis y label/.style={at=(current axis.above origin),anchor=south},
  every axis x label/.style={at=(current axis.right of origin),anchor=west},
  height=4.5cm, width=10cm,
  xtick=\empty,
  ytick=\empty,
  enlargelimits=false, clip=false, axis on top,
  grid = major
  ]
  \addplot [fill=red_1, draw=none, fill opacity = 1, domain=0:8] {7*gauss(4,1.5)} \closedcycle;
  \addplot [fill=blue_1, draw=none, fill opacity = 1, domain=0:4] {gauss(2,0.8)} \closedcycle;
  \addplot [fill=blue_1, draw=none, fill opacity = 1, domain=4:8] {gauss(5.8,0.5)} \closedcycle;
  \addplot [very thick,blue_1] {gauss(2,0.8)};
  \addplot [very thick,blue_1] {gauss(5.8,0.5)};
  \addplot [very thick,red_1] {7*gauss(4,1.5)};
  \addplot[thick, samples=50, dashed, domain=0:8, black] coordinates {(4,0)(4,2.2)};
\end{axis}

\node[rectangle, inner sep=0pt, minimum size=0.2cm, minimum height=0.1cm, fill=none] at (2.0, 1.7) (SYNTHETIC) {};

\node[rectangle, right of=SYNTHETIC, xshift=-0.5cm] (S) {\textbf{\textcolor{red_1}{$P_Z$}}};

\node[rectangle, inner sep=0pt, minimum size=0.2cm, minimum height=0.1cm, fill=none] at (0.2, 0.5) (ORIGINAL) {};

\node[rectangle, right of=ORIGINAL, xshift=-0.5cm] (O) {\textbf{\textcolor{blue_1}{$P$}}};

\node[rectangle, inner sep=0pt, minimum size=0.2cm, minimum height=0.1cm, fill=none, anchor=west] at (4.35, 2.65) (BOUNDARY) {$f_\mathcal{O}$};

\end{tikzpicture}
\caption{Gaussian training data distribution $P$ and learned decision boundary $f_\mathcal{O}$. Alternative gaussian distribution for $P_Z$.}\label{fig:distribution}
\end{figure}

Take for example the completely separable binary problem in Fig.~\ref{fig:distribution}, where each class comes from a Gaussian distribution and the decision boundary lies in a low density area of the space. Further assume that we are in a production setting, so that we have full knowledge of the system. In principle, in this scenario it would be possible, and even desirable, to match $P_Z$ with $P$. Indeed, by forcing $P_Z = P$ we ensure that the copy replicates the learned decision behaviour in those areas where the training data lie. However, the copy may display a completely different behaviour around the boundary, where these data are scarce. An interesting modelling question in this scenario would be: {\it what should the copy do in corner cases?} Another extreme case is that of counterfactuals, which include operation regimes even in front of impossible events and data values. 

\pgfmathdeclarefunction{gauss}{2}{%
  \pgfmathparse{1/(#2*sqrt(2*pi))*exp(-((x-#1)^2)/(2*#2^2))}%
}

More generally, defining $P_Z$ to resemble the form of $P$ might help in ensuring that the copy generalizes well in the training domain. However, this can also be achieved by other methods, such as updating the form of $P_Z$ as we gain more information about $f_\mathcal{O}$, or choosing a $P_Z$ that adapts to the form of the copy hypothesis space. Indeed, choosing $P_Z$ adequately can be difficult, given that we have no intuition about where the training data are located or which specific regions the copy should focus on. In Sec.~\ref{sec:insights} we study this problem in more depth. 

\subsection{Introducing the dual optimization}

Let us then assume an arbitrary form for the probability distribution $P_Z$. Because maximizing the posterior is equal to maximizing the log-posterior, we rewrite (\ref{eq:copy_inference_log_int}) as 
\begin{flalign*}
    \theta^* &= \arg\max_{\theta} \bigg[\log \bigg( \int_{\boldsymbol{z}\sim P_Z} P(\theta|f_{\mathcal{O}}(\boldsymbol{z})) dP_Z\bigg) \bigg] \\
    &= \arg\max_{\theta} \bigg[\log \bigg( \int_{\boldsymbol{z}\sim P_Z} \frac{P(f_\mathcal{O}(\boldsymbol{z})|\theta)P(\theta)}{P(f_\mathcal{O}(\boldsymbol{z}))} dP_Z\bigg) \bigg]
\end{flalign*}

\noindent
where we apply \textit{Bayes' rule} to the terms inside the integral. Using \textit{Jensen's inequality}\footnote{Jensen's inequality states that for any concave function $f$ it holds that $E[f(X)] \leq f(E[X])$. In particular, for the $\log(x)$ function.} we can then provide a lower bound for $\theta^*$ of the form\footnote{Maximization of the lower bound also maximizes the original function. However, the optimal value of the lower bound may differ from that of the original objective function.}

\begin{flalign}
    \theta^* &= \arg\max_{\theta} \int_{\boldsymbol{z}\sim P_Z} \log \bigg( \frac{P(f_\mathcal{O}(\boldsymbol{z})|\theta)P(\theta)}{P(f_\mathcal{O}(z))} \bigg) dP_Z \nonumber\\
    &= \arg\max_{\theta} \bigg[ \int_{\boldsymbol{z}\sim P_Z} \log P(f_\mathcal{O}(\boldsymbol{z})|\theta) dP_Z - \nonumber \\
    & \qquad \qquad \quad \ \ \int_{\boldsymbol{z}\sim P_Z} \log P(f_\mathcal{O}(\boldsymbol{z}))dP_Z + \log P(\theta) \bigg] \nonumber \\
    &= \arg\max_{\theta} \bigg[ \int_{\boldsymbol{z}\sim P_Z} \log P(f_\mathcal{O}(\boldsymbol{z})|\theta) dP_Z + \log P(\theta) \bigg] \label{eq: copy_parameters}
\end{flalign}
\noindent
where we drop the term $\int_{\boldsymbol{z}\sim P_Z} \log P(f_\mathcal{O}(\boldsymbol{z}))dP_Z$, which has no dependence on $\theta$.

The solution to (\ref{eq: copy_parameters}) depends on the form of the considered models. In this seminal article we study hard decision copies. Under this framework, we can recover regularized empirical risk minimization models \cite{Vapnik2000} if we approximate the distributions above with an exponential family

\begin{eqnarray*}
   P(f_{\mathcal{O}}(\boldsymbol{z})|\theta) \propto e^{-\gamma_1 \ell_1(f_{\mathcal{C}}(\boldsymbol{z},\theta),f_{\mathcal{O}}(\boldsymbol{z}))}; &  P(\theta)\propto e^{-\gamma_2 \ell_2(\theta, \theta^+)}\\
\end{eqnarray*}
\noindent
for $\ell_i(a,b)$ a measure of disagreement between $a$ and $b$, and $\theta^+$ our prior about $\theta$. Using this approximation we can rewrite (\ref{eq: copy_parameters}) as

\begin{flalign}
    \theta^* &= \arg\min_{\theta} \bigg[\int_{\boldsymbol{z}\sim P_Z} \gamma_1\ell_1(f_{\mathcal{C}}(\boldsymbol{z},\theta),f_{\mathcal{O}}(\boldsymbol{z})) dP_Z \nonumber\\
    & \qquad \qquad \quad + \gamma_2\ell_2(\theta,\theta^+) \bigg]
\label{eq:copy_inference_}
\end{flalign}

The first term in this expression is the expected value of the disagreement between model and copy, which has the form of empirical risk minimization. The expected loss particularized to our copying problem can be defined as

\begin{equation}
    R^\mathcal{F}(f_{\mathcal{C}}(\boldsymbol{z}, \theta),f_{\mathcal{O}}(\boldsymbol{z})) = \mathbb{E}_{\boldsymbol{z}\sim P_Z}[\ell_1(f_{\mathcal{C}}(\boldsymbol{z}, \theta), f_{\mathcal{O}}(\boldsymbol{z}))]
    \label{eq:expected_loss}
\end{equation}
\noindent
over the probability distribution $P_Z$. We refer to this value as the \textit{fidelity error}. This error captures all the loss of copying. In the general form, it corresponds to the integral $\int_{\boldsymbol{z}\sim P_Z} \log P(f_\mathcal{O}(z)|\theta) dP_Z$ in (\ref{eq: copy_parameters}), \textit{i.e.} the probability that the copy resembles the model.

The second term in (\ref{eq:copy_inference_}) refers to the fit of the parameters to the prior and can be identified as the regularization term

\begin{displaymath}
    \Omega(\theta) = \ell_2(\theta,\theta^+).
\end{displaymath}

Under the empirical risk minimization framework we approximate the expected loss by the empirical risk. The particularization of the empirical risk to the copying setting corresponds to the \textit{empirical fidelity error}, $R^{\mathcal{F}}_{emp}$. We define this value as the empirical version of the fidelity error

\begin{equation}
    R^{\mathcal{F}}_{emp}(f_{\mathcal{C}}(\boldsymbol{z}, \theta),f_{\mathcal{O}}(\boldsymbol{z})) = \frac{1}{N}\sum_{j=1}^N \ell_1(f_{\mathcal{C}}(\boldsymbol{z_j}, \theta), f_{\mathcal{O}}(\boldsymbol{z_j})) \label{emp_fid_error}
\end{equation}
\noindent
and rewrite (\ref{eq:copy_inference_}) for the discrete case as follows

\begin{flalign}
    (\theta^*,\boldsymbol{Z}^*) &= \arg\min_{\theta,\boldsymbol{z}\sim P_Z} \bigg[R^{\mathcal{F}}_{emp}(f_{\mathcal{C}}(\boldsymbol{z}, \theta),f_{\mathcal{O}}(\boldsymbol{z})) +  \Omega(\theta)\bigg] \nonumber\\
    &= \arg\min_{\theta,\boldsymbol{z}\sim P_Z} \bigg[\frac{1}{N}\sum_{j=1}^N \gamma_1\ell_1(f_{\mathcal{C}}(\boldsymbol{z}_j,\theta),f_{\mathcal{O}}(\boldsymbol{z}_j)) \nonumber\\
    & \qquad \qquad \qquad + \gamma_2\ell_2(\theta,\theta^+) \bigg],\label{eq:copy_inference_approx_}
\end{flalign}
\noindent
where $\boldsymbol{Z}$ corresponds to the set of synthetic samples $\boldsymbol{z} \sim P_Z$. We refer to the set of labelled synthetic pairs $\mathscr{Z} = \{(\boldsymbol{z}_j, f_\mathcal{O}(\boldsymbol{z}_j))\}^{N}_{j=1}$ as the \textit{synthetic dataset}. The expression above is a dual optimization, where we simultaneously optimize the copy parameters $\theta$ and the synthetic set $\mathscr{Z}$. This duality results from referring to the decision function $f_\mathcal{O}$ instead of exploiting the training data $\mathscr{D}$, and it fundamentally shapes how copying works.

\begin{figure*}[!t]
\centering
\includegraphics[width=16cm,trim={0 1cm 0 0},clip]{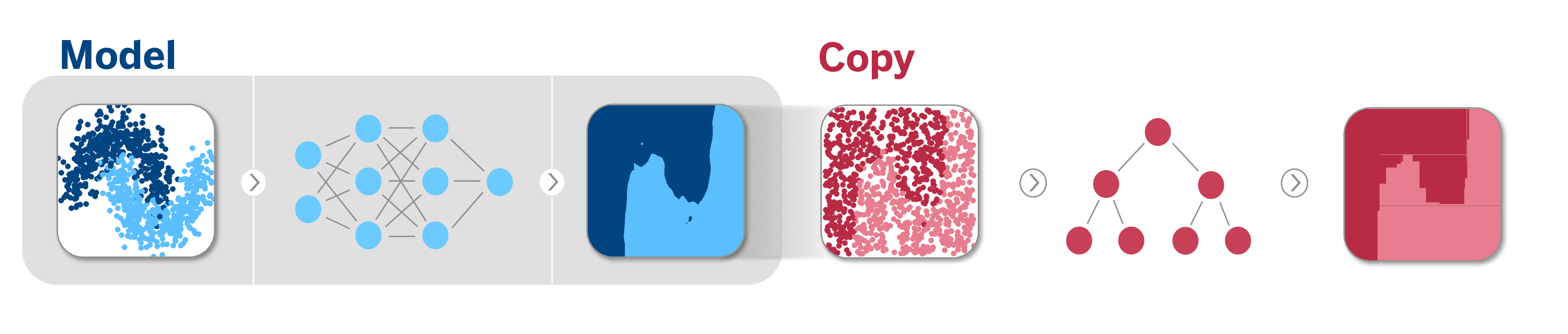}
\caption{Example of the \textit{single-pass copy}.
}\label{fig:pipeline}
\end{figure*}

\subsection{Why copying is not learning}

The class membership predictions of $f_\mathcal{O}$ define a hard classification boundary. The resulting problem has two important characteristics: (i) the synthetic dataset is always separable and (ii) a potentially infinite stream of synthetic data is accessible. These properties define copying as a problem different from learning, as traditionally understood by the machine learning community.

Because the synthetic set is separable, if we assume a copy with enough capacity, it is always possible to achieve zero empirical error,  $R_{emp}^{\mathcal{F}}(f_{\mathcal{C}}(\boldsymbol{z}, \theta),f_{\mathcal{O}}(\boldsymbol{z})) = 0$. The error then only depends on the generalization gap for the synthetic dataset. And since we can generate infinite synthetic data, this value can be asymptotically reduced to zero. Hence, in theory, copying can be performed without loss and redefined as the unconstrained optimization problem  

\begin{equation}\label{eq:unconstrained}
\underset{\theta,\boldsymbol{Z}}{\text{minimize}}  \quad
R^{\mathcal{F}}_{emp}(f_{\mathcal{C}}(\boldsymbol{z}, \theta),f_{\mathcal{O}}(\boldsymbol{z})).
\end{equation}

Yet, in practice, the synthetic set is finite. It therefore stands to reason to impose that the copy have small capacity, $\Omega(\theta)$, and rewrite the copying problem as 

\begin{flalign}\label{eq:capacity}
\underset{\theta,\boldsymbol{Z}}{\text{minimize}}  &\quad \Omega(\theta)\\
\text{subject to} & \quad \|R^{\mathcal{F}}_{emp}(f_{\mathcal{C}},f_{\mathcal{O}})-R^{\mathcal{F}}_{emp}(f^{\dagger}_{\mathcal{C}},f_{\mathcal{O}})\|<\epsilon, \nonumber
\end{flalign}
\noindent
for $f^{\dagger}_{\mathcal{C}}$ the solution to (\ref{eq:unconstrained}) and $\epsilon$ a defined tolerance\footnote{In what follows, we favour a more concise notation and drop the explicit dependence on the synthetic data $\boldsymbol{z}$ and copy parameters $\theta$.}. The term $\|R^{\mathcal{F}}_{emp}(f_{\mathcal{C}},f_{\mathcal{O}})-R^{\mathcal{F}}_{emp}(f^{\dagger}_{\mathcal{C}},f_{\mathcal{O}})\|<\epsilon$ defines a feasible set of parameters. The solution to (\ref{eq:capacity}) achieves the smallest capacity while keeping $R^{\mathcal{F}}_{emp}(f_{\mathcal{C}},f_{\mathcal{O}})$ within a tolerance of the unconstrained optimal value of the empirical fidelity error, $R^{\mathcal{F}}_{emp}(f^{\dagger}_{\mathcal{C}},f_{\mathcal{O}})$. We argue that there exists a set of parameters $\theta$ that fulfill this constraint.

In some cases the optimal loss value is known in advance. Consider, for example, the hinge-loss in SVMs, where $R^{\mathcal{F}}_{emp}(f^{\dagger}_{\mathcal{C}},f_{\mathcal{O}}) = 0$. However, this is not always the case, e.g. least-square errors in classification\footnote{Instead of tracking the empirical risk we can track the empirical error, which can be set to zero due to the separability property.}. Copying is different from the standard multi-objective optimization in a pure learning setting, where the optimal values of both the loss and the regularization term are unknown. Instead of having a \textit{Pareto's surface} of plausible optimal solutions, as long as $\Omega(\theta)$ is convex, the solution to (\ref{eq:capacity}) is unique.

This optimization can be straightforwardly solved in cases where the capacity is directly modelled, such as those of SVMs and neural networks, using a regularization function, or Bayesian models, selecting the priors. For other models, such as trees, the complexity control must be done by either early stopping or by an external process, such as post- or pre-pruning. Finally, techniques such as boosting or deep learning may exhibit a delayed overfitting effect \cite{Schapire1998BoostingMethods,Neyshabur2017Geometry,Brutzkus2017SGDData}. A property that can be exploited to our advantage to directly solve (\ref{eq:unconstrained}) instead of (\ref{eq:capacity}).

\subsection{The single-pass copy}

Conducting a simultaneous optimization of the synthetic data and the copy parameters requires the copy hypothesis space to have certain properties, such as online updating. This challenging issue is out of the scope of this paper and requires further research. Hence, for the sake of simplicity, in the rest of this article we consider the simplest approach to solving the dual copying problem: the {\it single-pass copy}. We cast the simultaneous optimization problem into one where only a single iteration of an alternating projection optimization scheme is used. This effectively splits the problem in two independent sub-problems:\\

\noindent
\textbf{Step 1: Synthetic sample generation}. The first step is to find the optimal set $\boldsymbol{Z}^*$. This set is that for which the empirical fidelity error, $R^{\mathcal{F}}_{emp}$, is minimal
    \begin{displaymath}
    \boldsymbol{Z}^* = \arg\min_{\boldsymbol{Z}} R^{\mathcal{F}}_{emp}
    \end{displaymath}
\noindent
As a result, we obtain the optimal synthetic dataset $\mathscr{Z}^*$.\\

\noindent
\textbf{Step 2: Building the copy}. Once having generated  and labelled the set $\mathscr{Z}^*$, the next step is to find $\theta^*$ such that
    \begin{flalign*}
    \underset{\theta}{\text{minimize}}  &\quad \Omega(\theta)\\
    \text{subject to} & \quad \|R^{\mathcal{F}}_{emp}(f_{\mathcal{C}},f_{\mathcal{O}})-R^{\mathcal{F}}_{emp}(f^{\dagger}_{\mathcal{C}},f_{\mathcal{O}})\|<\epsilon, \nonumber
    \end{flalign*}
\noindent    
or its simplified version (\ref{eq:unconstrained}), provided that the adequate conditions hold.\\

An example of the single-pass copy is shown in Fig.~\ref{fig:pipeline}, where the binary decision function learned by a fully-connected neural network is copied with a decision tree classifier. The tree-based copy is built using a set of synthetic samples drawn from a uniform distribution and labelled according to the hard predictions output by the neural net. 

\section{Meaningful Insights}
\label{sec:insights}

In what follows, we bridge the gap between theory and practice by using toy problems to draw relevant conclusions from the  derivation above. We focus on the two steps of the single-pass copy: we begin by studying the synthetic sample generation process and then show how copying differs from learning in a practical setting. 

\subsection{STEP 1: Synthetic sample generation }

\begin{figure*}[!t]
\centering
\begin{tabular}{ccc}
\adjustbox{valign=t}{\includegraphics[height=1.0in]{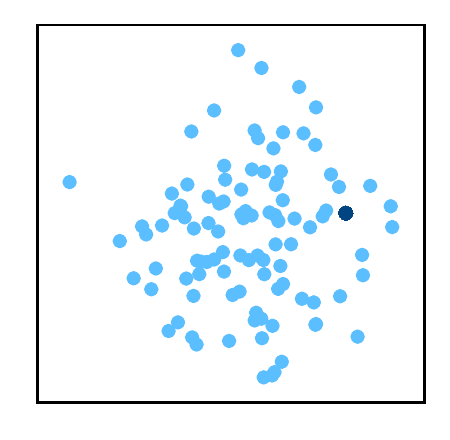}}\label{fig:trainingdata} & \multirow{3}{*}{\adjustbox{valign=t}{\includegraphics[height=2.17in]{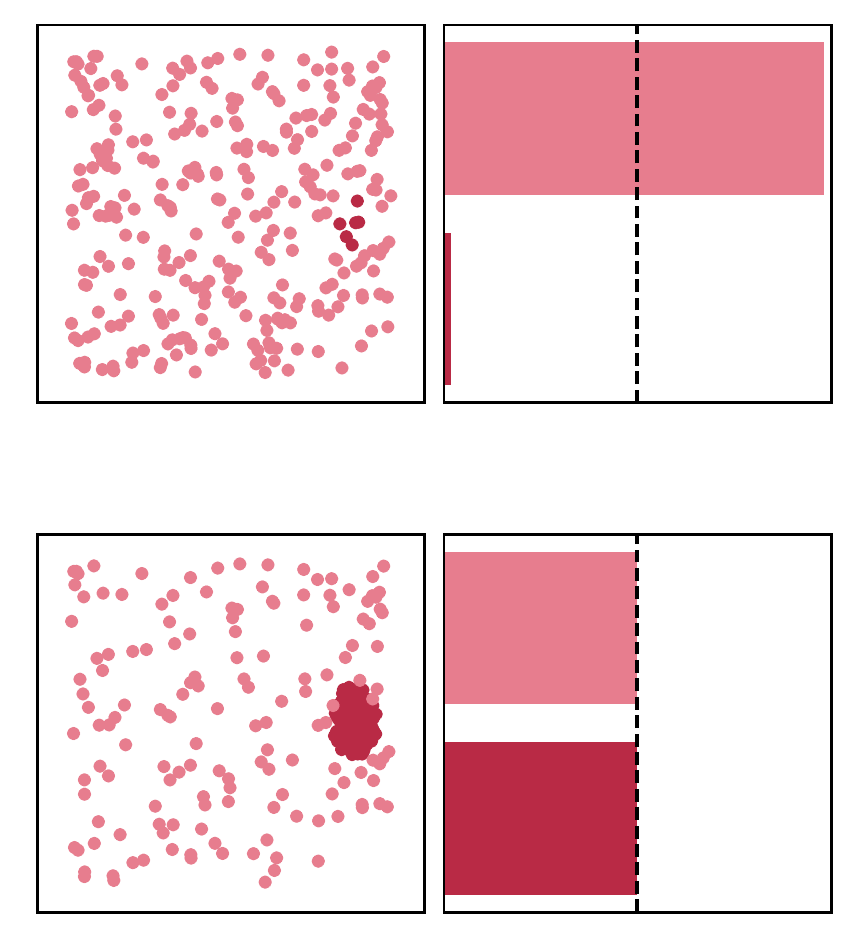}}\label{fig:uniform}}
 &
\multirow{3}{*}{\adjustbox{valign=t}{\includegraphics[height=2.17in]{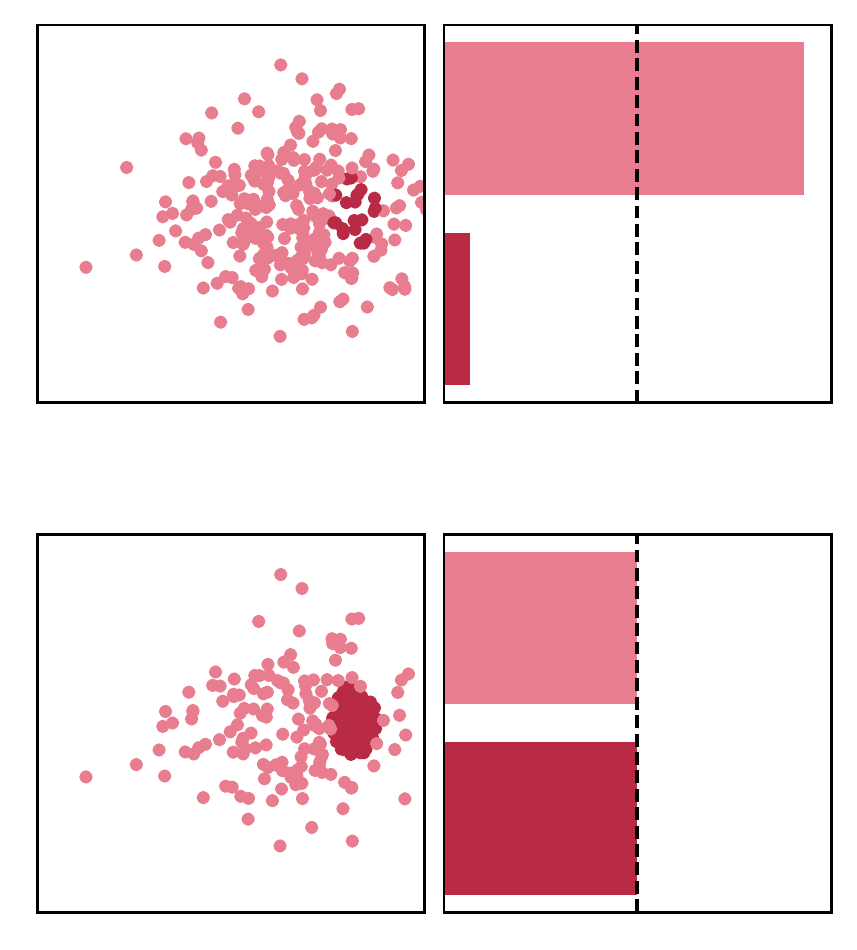}}\label{fig:normal}} \\
 \sffamily\sansmath\footnotesize{(a)} & & \\
\includegraphics[height=1.0in]{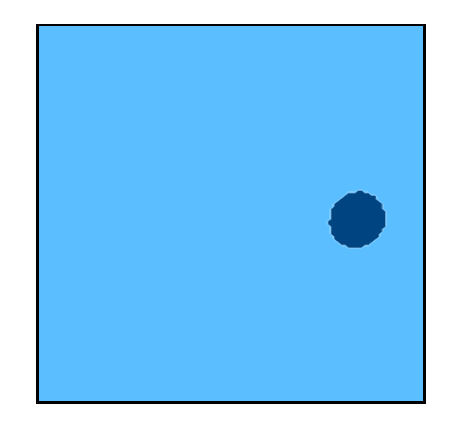}\label{fig:model} & &
 \\
 \sffamily\sansmath\footnotesize{(b)} &  \sffamily\sansmath\footnotesize{(c)} &  \sffamily\sansmath\footnotesize{(d)} \\
\end{tabular}
\caption{(a) Training dataset. (b) Decision boundary learned by a Gaussian Process classifier. From top to bottom raw and balanced synthetic datasets generated from (c) a uniform distribution and (d) a standard normal distribution.}
\label{fig:sampling}
\end{figure*}

For the sake of this discussion, let us consider a binary classification problem and let $f_{\mathcal{O}}(\boldsymbol{z})\in\{-1,+1\}$ and $f_{\mathcal{C}}(\boldsymbol{z}, \theta)\in\{-1,+1\}$, for any $\boldsymbol{z} \in \mathcal{X}$. Let us also consider the case where $\ell_1$ corresponds to the $0/1$ loss. For this case, the empirical fidelity error in (\ref{emp_fid_error}) can be rewritten as

\begin{flalign*}
    R^{\mathcal{F}}_{emp} &= \frac{1}{2N}\sum_{j=1}^N \Big| f_{\mathcal{O}}(\boldsymbol{z_j})) - f_{\mathcal{C}}(\boldsymbol{z_j}, \theta) \Big| \\
    &= \frac{1}{2N}\sum_{j=1}^N \Big| f_{\mathcal{O}}(\boldsymbol{z}) \Big| \Big| 1-\frac{f_{\mathcal{C}}(\boldsymbol{z}, \theta)}{f_{\mathcal{O}}(\boldsymbol{z})} \Big| \\
    & = \frac{1}{2N}\sum_{j=1}^N \Big( 1-\frac{f_{\mathcal{C}}(\boldsymbol{z}, \theta)}{f_{\mathcal{O}}(\boldsymbol{z})} \Big) \\
    & = \frac{1}{2N}\sum_{j=1}^N  1 - \frac{1}{2N}\sum_{j=1}^N \frac{f_{\mathcal{C}}(\boldsymbol{z}, \theta)}{f_{\mathcal{O}}(\boldsymbol{z})} \\
    & = \frac{1}{2} - \frac{1}{2N}\sum_{j=1}^N f_{\mathcal{C}}(\boldsymbol{z}, \theta)f_{\mathcal{O}}(\boldsymbol{z}), \; \boldsymbol{z}^{(N)}\sim P_Z 
\end{flalign*}

Let us now define a partition of the space such that $\mathcal{X}=\mathcal{X_+} \cup \mathcal{X_-}$ and $\mathcal{X_+} \cap \mathcal{X_-}=\emptyset$, where $\mathcal{X_+}=\{\boldsymbol{z}|\boldsymbol{z}\in \mathcal{X}, f_{\mathcal{O}}(\boldsymbol{z})=1\}$ and 
 $\mathcal{X_-}=\{\boldsymbol{z}|\boldsymbol{z}\in \mathcal{X}, f_{\mathcal{O}}(\boldsymbol{z})=-1\}$ are the two sub-spaces defined by the model. We rewrite the equation above in terms of this partition as

\begin{displaymath}
R^{\mathcal{F}}_{emp} =  \frac{1}{2} -\frac{1}{2N_+}\sum_{j=1}^{N_+}f_{\mathcal{C}}(\boldsymbol{z}_j, \theta) +\frac{1}{2N_-}\sum_{j=1}^{N_-} f_{\mathcal{C}}(\boldsymbol{z}_j, \theta)
\end{displaymath}
\noindent
for $N_+$ and $N_-$ the number of samples lying in $\mathcal{X_+}$ and $\mathcal{X_-}$, respectively. 

We define the probability of a sample lying in $\mathcal{X_+}$ as $p_+ = \mathbb{P}(\boldsymbol{z}\in \mathcal{X_+})$ and the probability of a sample lying in $\mathcal{X_-}$ as $p_- = \mathbb{P}(\boldsymbol{z}\in \mathcal{X_-})$. These two probabilities depend on the \textit{size} of the positive and negative domains. In particular,

\begin{displaymath}
    p_+  = \int_{\boldsymbol{z}\in \mathcal{X_+}} P_Z(\boldsymbol{z}) dz,  \quad p_- = \int_{\boldsymbol{z}\in \mathcal{X_-}} P_Z(\boldsymbol{z}) dz.
\end{displaymath}

With these quantities, we can see that $N_+=N p_+$ and  $N_-=N p_-$. Thus,

\begin{displaymath}
R^{\mathcal{F}}_{emp} = \frac{1}{2} -\frac{1}{2Np_+}\sum_{j=1}^{Np_+} f_{\mathcal{C}}(\boldsymbol{z}_j, \theta) +\frac{1}{2Np_-}\sum_{j=1}^{Np_-} f_{\mathcal{C}}(\boldsymbol{z}_j, \theta).
\end{displaymath}

Minimization of this expression explicitly depends on the form of $P_Z$. In the simplest case, we can assume this distribution to be flat on the domain $\mathcal{X}$, so that $\boldsymbol{z}\sim \mathcal{U}(\mathcal{X})$. Under this assumption, $p_+$ and $p_-$ correspond to the fraction of volume for each of the classes. Recalling the form of the error for the Monte Carlo estimator under this distribution, we can express the standard error for $R^{\mathcal{F}}_{emp}$ as

\begin{displaymath}
\sigma(R_{\mathcal{CV}}) \propto \mathcal{O}\Big(\frac{1}{\sqrt{N p_+}}+ \frac{1}{\sqrt{N p_-}}\Big).
\end{displaymath}

We exploit this expression to extract relevant insights for the synthetic sample generation process. First, we confirm the need to define an attribute representation $\mathcal{X}$. This is a reasonable assumption, since we need to have an approximate idea of the dynamic range of all variables in order to build meaningful queries. 

Second, we note that in some situations there might be a mismatch between the decision boundary achievable by the copy and $f_\mathcal{O}$. As a consequence, a given synthetic dataset may not perform equally for different copy hypotheses. Consider a non-linear decision function and a linear copy model. Exploring the twists of the decision boundary during the synthetic sample generation process may not be relevant in this situation. Thus, we should consider the properties and assumptions of the copy hypothesis space to effectively exploit each generated sample. 

Another important issue is that of volume imbalance, which arises when one or more of the classes occupy a region of the space much smaller than the rest. 

\subsubsection{The issue of volume imbalance}

The empirical fidelity error depends on the fraction of volume occupied by each decision region. If the spatial support of one class is small with respect to the total volume, it may be difficult to have a meaningful number of samples on that region, resulting in large approximation errors.  

In Fig.~\ref{fig:sampling}(a), we show a binary dataset with a balanced label distribution. Despite the number of instances per label being equal, note that there are notable differences in the volume of each of the classes. The resulting decision function is displayed in Fig.~\ref{fig:sampling}(b).

To copy this model, we assay two different forms for $P_Z$. In a preliminary approach, we generate samples at random until we reach a desired number of points. In Fig.~\ref{fig:sampling}(c) and Fig.~\ref{fig:sampling}(d) we plot the sets that result for a uniform distribution and for a standard normal distribution, respectively. The resulting data, shown together with their corresponding label distribution, are notably imbalanced: there is one class for which we only recover a few number points. This result is unrelated to class distribution.

Fortunately, the volume imbalance effect can be alleviated either by a good choice of $P_Z$ or by imposing that the resulting set be balanced. For example, we can try to infer a sampling distribution that allocates a large amount of the probability mass around the unknown decision boundary. Due to its complexity, we believe the problem of finding an optimal $P_Z$ to be out of the scope of this work. This issue will be subject to further analysis in future contributions. Indeed, in a recent paper \cite{diego} we have studied different sampling algorithms for the copying setting, including a technique that focuses on boundary exploration, a Bayesian-based optimizer, a modified version of the Jacobian approach proposed by \cite{Papernot2017PracticalLearning} and raw random sampling.

Alternatively, we can overcome the issue of volume imbalance using heuristics that balance a general exploration of the space with exploitation around the areas of interest. Hence,  we impose that the resulting set be balanced with respect to the class labels. We force the data generator to focus on those areas where the misrepresented class is located, to ensure that all labels are well represented in the resulting set, as shown in Fig.~\ref{fig:sampling}. 

\subsection{STEP 2: Building the copy}

\begin{figure}[!t]
\centering
\begin{tabular}{c c}
     \adjustbox{valign=t}{\includegraphics[width=0.9in, trim={0.2in 0 0 0 }]{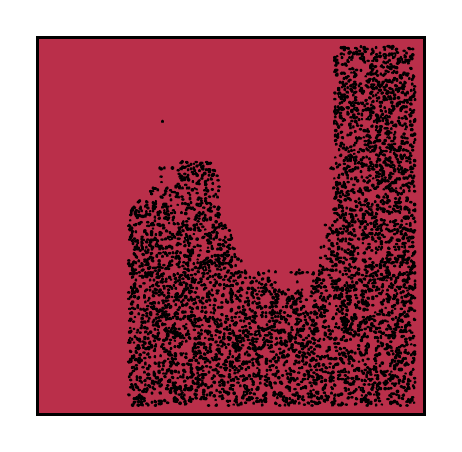}} & \multirow{5}{2.0in}{\adjustbox{valign=t}{\includegraphics[width=2.1in, trim={0.2in 0 0 0.1in}]{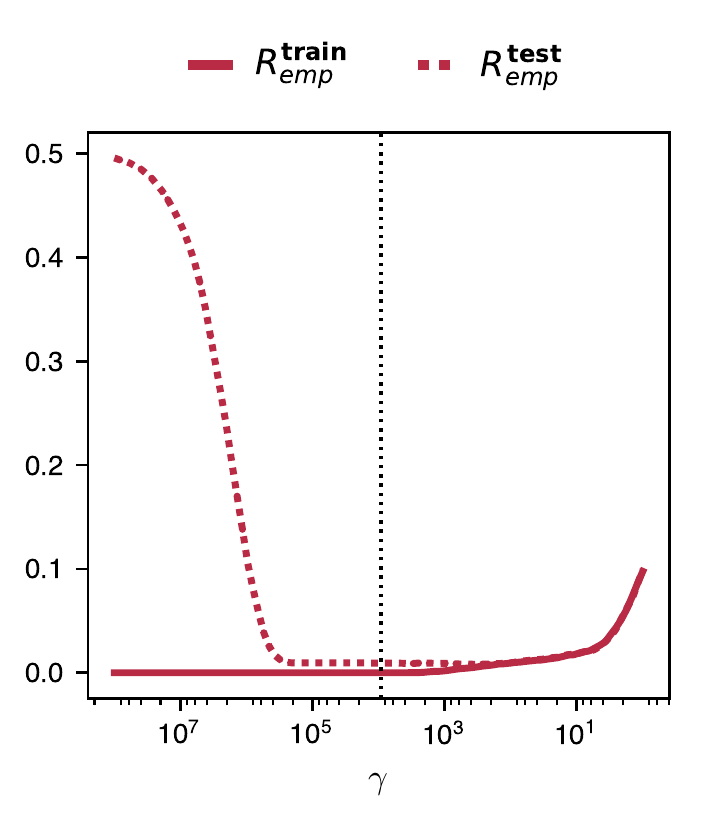}}}\\
     \sffamily\sansmath\footnotesize{(a)} &  \\
     & \\
     \includegraphics[width=0.9in, trim={0.2in 0 0 0}]{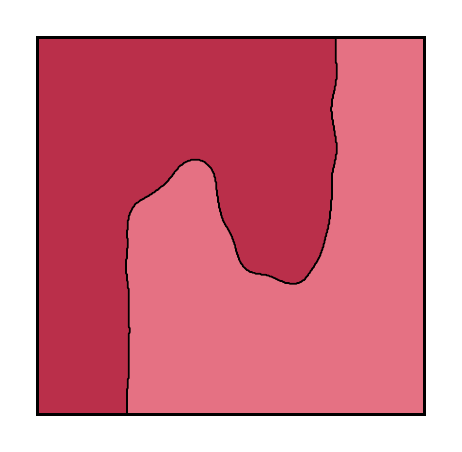} &  \\
      & \\
     \sffamily\sansmath\footnotesize{(b)} & \sffamily\sansmath\footnotesize{(c)} \\
\end{tabular}
\caption{Decision boundaries learned by copies with (a) a maximal and (b) an optimal $\gamma$. (c) Empirical risk and generalization error for decreasing values of $\gamma$. The dotted black line refers to the optimal $\gamma$.}
\label{make_moons.copy}
\end{figure}

The second part of the alternating projection scheme corresponds to finding the optimal parameters for the copy. For illustration purposes, consider a radial basis function kernel SVM. This model is defined by a kernel function of the form $\mathcal{K}(\boldsymbol{x},\boldsymbol{x}') = e^{-\gamma ||\boldsymbol{x}-\boldsymbol{x}'||^2}$, where $||\boldsymbol{x}-\boldsymbol{x}'||^2$ corresponds to the squared Euclidean distance, and $\gamma$ is the inverse of the radius of influence of the support vectors, i.e. the width of the kernel. This means, in essence, that $\gamma$ controls the capacity: the larger its value, the higher the complexity. In other words, minimizing the model capacity in (\ref{eq:capacity}) amounts to minimizing $\gamma$. In Fig.~\ref{make_moons.copy} we show how this can be exploited in practice to copy the neural net in Fig.~\ref{fig:pipeline} using synthetic samples drawn at random from a uniform distribution. 

In particular, Fig.~\ref{make_moons.copy}(a) shows the copy decision function for a maximal value of $\gamma$, such that the second term in (\ref{eq:capacity}) is satisfied and the empirical error is zero. Fig.~\ref{make_moons.copy}(b) shows the decision boundary for a copy with optimal capacity $\gamma$, computed for a tolerance $\epsilon = 1e-4$. This solution results from sequentially reducing the value of $\gamma$ and monitoring the change in accuracy until the error deviation is greater than $\epsilon$. When comparing both plots we observe the improvement in generalization performance. This improvement is also seen in Fig.~\ref{make_moons.copy}(c), where train and generalization errors of the copy are shown for decreasing values of $\gamma$. For a bounded value of the empirical error, the generalization error is reduced as we decrease the capacity of the copy.

Unlike the classical machine learning, where capacity is optimized during the validation step, this result shows that it is possible to optimize the capacity of a copy during training. This has a profound impact on how copying is performed and shows that copying is not learning, in the traditional meaning of the word.

\subsubsection{Capacity error}

Lastly, note that the specific choice of copy hypothesis has a significant impact on performance. Different capacity copies may behave very differently when confronted with the same set of synthetic data points.

We refer to the capacity of a classifier as a measure of its complexity. A mismatch of capacity between model and copy can lead to poor performance results, even in cases where the synthetic dataset properly covers the input space. Take for example the case of a linear logistic regression and a support vector machine. The decision functions resulting from building copies based on these two architectures are notably different. Given the same set of synthetic points, the logistic model may not able to fully recover the form of the considered decision boundary if this is non-linear. This is because the original classifier, is not contained in the new hypothesis space. In the case of the SVM, the mismatch in capacity is presumably not so pronounced and therefore the copy decision boundary may be much more precise.

\section{Empirical Validation}
\label{sec:experiments}

In this section we present our experiments to empirically validate copies in a variety of well-known problems that include a diverse selection of UCI datasets with different number of classes and dimensions. We begin by proposing a set of performance metrics.

\subsection{Performance metrics}

When evaluating copies, we may ask questions of the form: \emph{"what does the performance on a synthetic validation set tell us about the generalization of the copy?"}, \emph{"does the copy have enough capacity to replicate the decision function?"} or, more generally,  \emph{"what metrics should we use to evaluate copies in terms of the available information?"}. In what follows we introduce a set of definitions aimed at answering these questions.

\subsubsection{Empirical fidelity error}

We particularize the empirical fidelity error in (\ref{emp_fid_error}) to the $0/1$ loss and measure it over the synthetic set $\mathscr{Z}$ as
\begin{equation}
    R^{\mathcal{F}, \mathscr{Z}}_{emp} = \frac{1}{N}\sum_{j = 1}^{N} \mathds{I}[f_{\mathcal{O}}(\boldsymbol{z}_j)\neq f_{\mathcal{C}}(\boldsymbol{z}_j)]
\end{equation}
\noindent
for $\mathds{I}$ the indicator function. In resorting to Monte Carlo integration we here necessarily incur in an approximation error that depends, among other things, on the quality of the set $\mathscr{Z}$. As a result, a low $R^{\mathcal{F}, \mathscr{Z}}_{emp}$ is no absolute guarantee of a good copy. For this value to be a valid assessment of the total error, the synthetic dataset must be large enough to ensure coverage of the input space and the volume imbalance effect needs to be controlled for. 

In cases where the constraints of the copying scenario are relaxed and the training data $\mathscr{D}$ is accessible, we could also evaluate the empirical fidelity error over this set as

\begin{equation}
    R^{\mathcal{F}, \mathscr{D}}_{emp} = \frac{1}{M}\sum_{i = 1}^{M} \mathds{I}[f_{\mathcal{O}}(\boldsymbol{x}_i)\neq f_{\mathcal{C}}(\boldsymbol{x}_i)]
\end{equation}

For validation purposes, in the following we assume these data to be known. In general, $R^{\mathcal{F}, \mathscr{D}}_{emp}$ and $R^{\mathcal{F}, \mathscr{Z}}_{emp}$ yield very different values. This difference arises from the mismatch between the probability density functions $P$ and $P_Z$. 

\subsubsection{Copy accuracy}

To evaluate the copy generalization performance over $\mathscr{D}$ we introduce the {\it copy accuracy}, $\mathcal{A}_\mathcal{C}$, as follows

\begin{equation}
    \mathcal{A}_\mathcal{C} = \frac{1}{M}\sum_{i = 1}^{M} \mathds{I}[t_i= f_{\mathcal{C}}(\boldsymbol{x}_i)],
\end{equation}

\begin{figure*}[!ht]
\centering
\begin{tabular}{ccc}
\includegraphics[width=1.66in]{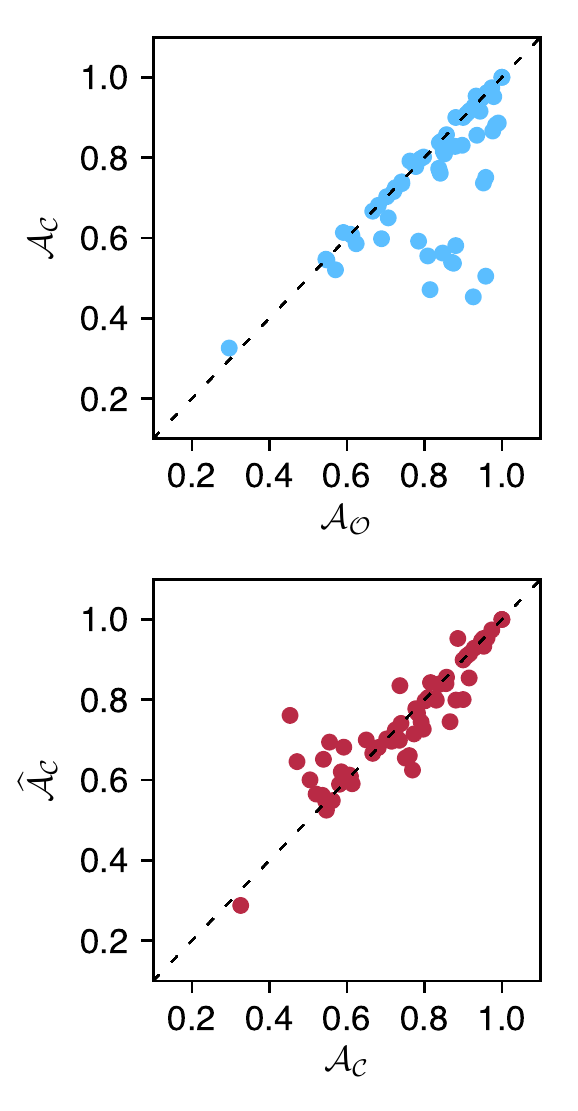}\label{fig:UCI_1} &
\includegraphics[width=1.5in]{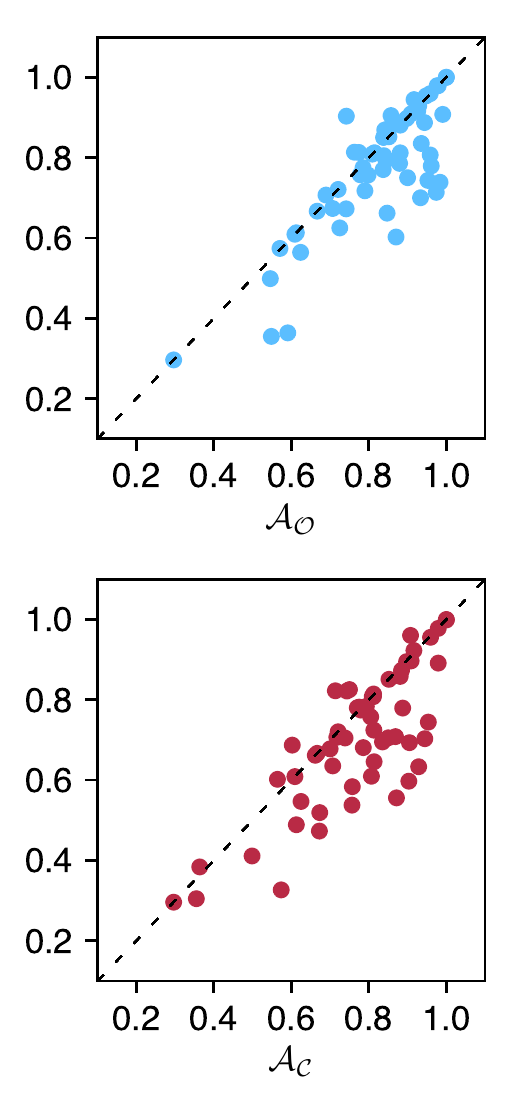}\label{fig:UCI_2} &
\includegraphics[width=1.5in]{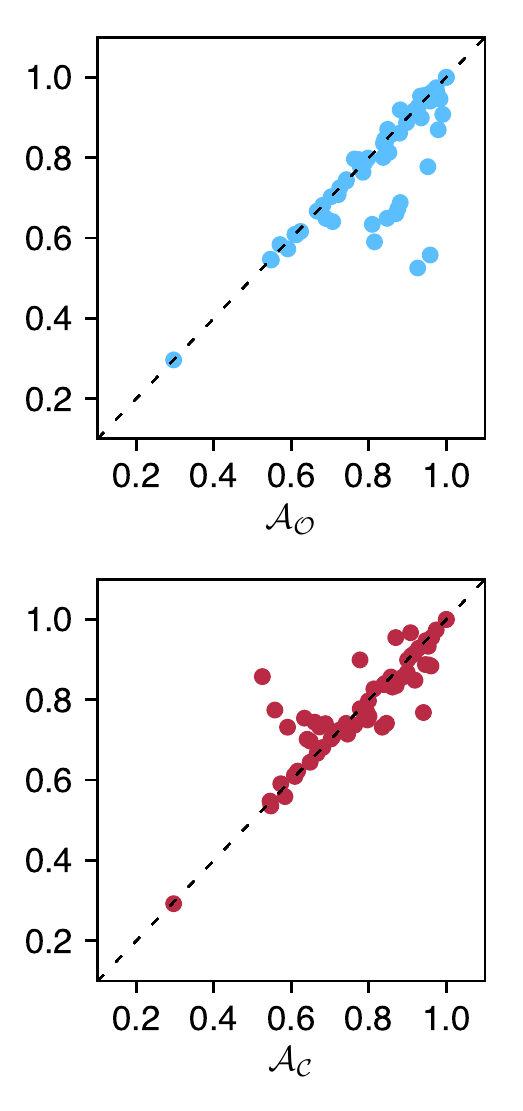}\label{fig:UCI_3} \\
\sffamily\sansmath\footnotesize{(a)} & \sffamily\sansmath\footnotesize{(b)} & \sffamily\sansmath\footnotesize{(c)} \\
\end{tabular}
\caption{From top to bottom, distribution of average copy accuracy against original accuracy and distribution of average estimated copy accuracy against average true copy accuracy for all datasets and for copies based on (a) decision trees, (b) logistic regression and (c) random forest.}
\label{fig:UCI}
\end{figure*}

\noindent
for $t \in \mathscr{T}$ the true labels. The performance of the copy on $\mathscr{D}$ is bounded by $\mathcal{A}_{\mathcal{O}}$, the accuracy of $f_\mathcal{O}$ on these data. In the ideal case the fidelity error is zero, so that $\mathcal{A}_{\mathcal{C}}= \mathcal{A}_{\mathcal{O}}$. In general, we can use the empirical fidelity error over the synthetic set to approximate $\mathcal{A}_{\mathcal{C}}$ by means of the \textit{estimated copy accuracy}, $\widehat{\mathcal{A}_{\mathcal{C}}}$, as follows 

\begin{equation}
    \widehat{\mathcal{A}_{\mathcal{C}}} = \mathcal{A}_{\mathcal{O}}(1-R^{\mathcal{F},\mathscr{Z}}_{emp}) \label{eq:estimated}
\end{equation}

\subsection{Experiments}

We use 60 datasets from the UCI Machine Learning Repository database \cite{Dheeru2017UCIRepository}. We refer the reader to \cite{Fernandez-Delgado2014DoProblems} for a specific description of initial data selection and preprocessing. We select those datasets with more than 100 samples and a frequency above $10\%$ for all class labels. We also require the number of inputs to be greater than double the number of attributes. Among the selected datasets 42 correspond to binary classification problems and 18 are multiclass. 

\subsubsection{Experimental set up}

We convert nominal attributes to numerical and re-scale variables to zero mean and unit variance. We split data into stratified 80/20 training and test sets. We use 6 state-of-the-art classification algorithms, including adaboost (\textit{\textit{adaboost}}), an artificial neural network (\textit{\textit{ann}}), a random forest (\textit{random\_forest}), a linear SVM (\textit{linear\_svm}), a SVM with a radial basis function kernel (\textit{rbf\_svm}) and a gradient-boosted tree (\textit{xgboost}). To avoid bias regarding the algorithm choice, we sort datasets in alphabetical order, group them in sets of 10 and randomly assign a classifier to each group.

We build a generic pipeline and train all models using a cross-validated grid-search over a fixed parameter grid. Three classifiers learn decision functions that exclude at least one of the class labels. This occurs for \textit{pittsburg-bridges-REL-L}, for which only two of the three classes are learned, and \textit{planning} and \textit{statlog-australian-credit}, for which a single class label is assigned to all data points. Besides, because we use a fixed pipeline, not all models yield an optimal performance. See, for example, the case of \textit{echocardiogram}, where accuracy is equal to $0.3$. 

We keep this result for two reasons. First, we want the experimental setup to be as agnostic as possible and hence the random pairing of models and datasets. Second, it reinforces an important idea: a copy can only be as good as the model it aims to replicate. Or in the other words, the baseline for the copy performance is the original model performance. Non-optimal models lead to poorly performing copies. We stress, nonetheless, that in a real setting one would be interested in copying only those models that perform reasonably well.

We draw $1e6$ random samples from a uniform distribution to generate balanced synthetic sets. We identify three cases of volume imbalance: \textit{congressional-voting}, \textit{ilpd-indian-liver} and \textit{statlog-image}. Despite the training data being balanced with respect to class distribution, we only recover a small fraction of samples for one or more of the labels. As previously mentioned, this could lead to sub-optimal results, given that the copy tends to wrongly classify points that belong to the subsampled classes. Imposing that the synthetic dataset be balanced mitigates this issue to a great extent and ensures that the copy treats all labels equally.

To evaluate the impact of heuristics, we assay different copy model hypotheses. We use decision trees because they are easily interpretable, logistic regression because it is a linear model and random forest as an example of a bagging method. We copy using no cross-validation or hyper-parameter tuning: trees are grown until each leaf contains a single sample and neural networks and boosting methods are trained with no regard for generalization. For validation purposes, we run each experiment 100 times and report averages over all repetitions for the true and the estimated copy accuracy. We also report the mean empirical fidelity error measured over both training and synthetic data.


\subsubsection{Results}

The measured performance metrics are shown in Fig.~\ref{fig:UCI}. In particular, Fig.~\ref{fig:UCI}(a), Fig.~\ref{fig:UCI}(b) and Fig.~\ref{fig:UCI}(c) show the distribution of the mean copy accuracy $\mathcal{A}_{\mathcal{C}}$ against the original accuracy $\mathcal{A}_{\mathcal{O}}$ and the estimated copy accuracy $\widehat{\mathcal{A}_{\mathcal{C}}}$ for all datasets  and copies based on decision trees (\textit{decision\_tree}), logistic regression (\textit{logistic\_regression}) and random forest (\textit{random\_forest}) classifiers, respectively. 

Results for both \textit{decision\_tree} and \textit{random\_forest} are scattered around the main diagonal, whereas copies based on \textit{logistic\_regression} show a greater dispersion; especially when comparing $\mathcal{A}_{\mathcal{C}}$ to $\widehat{\mathcal{A}_{\mathcal{C}}}$. In general, the value of $\widehat{\mathcal{A}_{\mathcal{C}}}$ is smaller than $\mathcal{A}_{\mathcal{C}}$, which means that the empirical fidelity error over the synthetic data overestimates the real error. This is in part due to the difference in the distributions $P$ and $P_Z$. When evaluating $R_{\mathcal{F}}^{\mathscr{Z}}$, we measure the performance of the copy in the space defined by $P_Z$, so that we may penalize the copy for errors in regions where there are no actual training data.

The complete summary of results for all problems and copy algorithms is shown in Table 3 in the Appendix. In most problems, results show the ability of copies to replicate the target decision behaviour. Overall, copy accuracy is competitive for the proposed synthetic dataset size and the estimated copy accuracy provides a reliable approximation to the accuracy of the copy in real data. The empirical fidelity error generally yields values close to 0, which indicates that copies are correctly built.

Table~\ref{tab:UCI_some} shows a selected set of results. There are several datasets where there is no degradation when using a \textit{logistic\_regression} to copy higher capacity models such as \textit{ann} or \textit{xgboost}. This is the case, for example, with \textit{breast-cancer-wisc} and \textit{wine}, where $\mathcal{A}_\mathcal{C}$ is reasonably close to $\mathcal{A}_\mathcal{O}$, even while the logistic model can only learn linear relations among attributes. We take this as an indication that the initial classifiers were too complex for the relatively simple problems. Copying here allows us to move to a more suitable solution, with less parameters and training requirements.

On the other hand, we identify a number of cases where copies based on \textit{decision\_tree} and \textit{random\_forest} clearly outperform \textit{logistic\_regression}. See, for example, \textit{energy-y1} and \textit{iris}. This is because when the decision function is not linear\footnote[6]{Despite the training data being linearly separable, the learned decision boundary may be non-linear.}, non-linear copies are needed. Here, the error due to a mismatch of capacity dominates, because the copy hypothesis space, the logistic family, does not contain $f_\mathcal{O}$.

Finally, in some instances the copy hypothesis space is well chosen and yet the empirical fidelity error is high. See for example \textit{musk\_1} and \textit{musk\_2}, which are both high dimensional problems where a \textit{linear\_svm} is copied using a \textit{random\_forest}. In both cases, $\mathcal{A}_\mathcal{C}$ is notably lower than $\mathcal{A}_\mathcal{O}$. This happens in complex datasets, where $1e6$ synthetic data points are probably not enough to ensure a small $R^\mathcal{F}_{emp}$.

\begin{table*}[!ht]
\centering
\caption{Subset of Relevant Results for the UCI Experiment. (*) Multiclass. ($\dagger$) More than 10 dimensions. ($\S$) Standard deviation above $0.05$.} \label{tab:UCI_some}
\small
    \begin{tabular}{@{}lccccccccccc} 
    \toprule
    \\[-1em]
    \multirow{3}{2.5cm}{Dataset} &
    \multirow{3}{0.8cm}{\centering{$\mathcal{H}_\mathcal{O}$}} & 
    \multirow{3}{0.6cm}{\centering{$\mathcal{A}_{\mathcal{O}}$}}&
    \multicolumn{3}{c}{\multirow{1}{*}{\centering{\textit{decision\_tree}}}} & \multicolumn{3}{c}{\multirow{1}{*}{\centering{\textit{logistic\_regression}}}} &
    \multicolumn{3}{c}{\multirow{1}{*}{\centering{\textit{random\_forest}}}}\\
    \\[-0.5em]
    \cline{4-12}
    \\[-0.75em]
    & & & \multirow{1}{0.6cm}{\centering{$\mathcal{A}_\mathcal{C}$}} & \multirow{1}{0.6cm}{\centering{$\widehat{\mathcal{A}}_\mathcal{C}$}} & \multirow{1}{0.6cm}{\centering{$R^{\mathcal{F},\mathscr{D}}_{emp}$}} & \multirow{1}{0.6cm}{\centering{$\mathcal{A}_\mathcal{C}$}} & \multirow{1}{0.6cm}{\centering{$\widehat{\mathcal{A}}_\mathcal{C}$}} & \multirow{1}{0.6cm}{\centering{$R^{\mathcal{F},\mathscr{D}}_{emp}$}} &
    \multirow{1}{0.6cm}{\centering{$\mathcal{A}_\mathcal{C}$}} & \multirow{1}{0.6cm}{\centering{$\widehat{\mathcal{A}}_\mathcal{C}$}} & \multirow{1}{0.6cm}{\centering{$R^{\mathcal{F},\mathscr{D}}_{emp}$}} \\
    \\[-0.5em]
    \toprule
    \\[-0.75em]
breast-cancer-wisc 	&	\textit{adaboost} 	&	\textbf{0.93}	&	 0.93 &	 0.9286 	&	 0.00 	&	 \textbf{0.93} 	&	 0.6333 	&	 0.06 	&	 0.93	&	 0.9286 	&	 0.00 \\
chess-krvkp$^\dagger$ 	&	\textit{ann} 	&	0.99	&	 0.89 &	 \textbf{0.9527}	&	 0.11 	&	 0.91 	&	 0.9603	&	 0.10 	&	 0.91 	&	 0.9670 	&	 0.09 \\
echocardiogram 	&	\textit{ann} 	&	\textbf{0.3}	&	 0.33 &	 0.2879	&	 0.05 	&	 0.30	&	 0.2960 	&	 0.00 	&	 0.30 	&	 0.2922 	&	 0.00 \\
energy-y1* 	&	\textit{ann} 	&	\textbf{0.96}	&	 \textbf{0.96} 	&	 0.9537 	&	 0.00 	&	 \textbf{0.78} 	&	 0.7744 	&	 0.23 	&	 \textbf{0.96} 	&	 0.9551 	&	 0.00\\
iris* 	&	\textit{random\_forest} 	&	0.93	&	 \textbf{0.95} 	&	 0.9332	&	 0.02	&	 \textbf{0.70} 	&	 0.6778 	&	 0.30	&	 \textbf{0.95} 	&	 0.9333 	&	 0.02 \\
musk-1$^\dagger$	&	\textit{linear\_svm} 	&	\textbf{0.88}	&	 0.54 	&	 0.5620 	&	 0.46$^\S$ 	&	 0.88 	&	 0.8732  	&	 0.01	&	 \textbf{0.67$^\S$} 	&	 0.7323$^\S$	&	 0.32\\
musk-2$^\dagger$ 	&	\textit{linear\_svm} 	&	\textbf{0.96}	&	 0.50$^\S$	&	 0.6005 	&	 0.50$^\S$	&	 0.96 	&	 0.9556 	&	 0.00 	&	 \textbf{0.56} 	&	 0.7745 	&	 0.44 \\
oocytes\_me\_nu\_4d$^\dagger$ 	&	\textit{linear\_svm} 	&	0.82	&	 0.47$^\S$	&	 0.6460 	&	 \textbf{0.52$^\S$}	&	 0.81 	&	 0.8144 	&	 0.00 	&	 0.59 	&	 0.7317 	&	 \textbf{0.38} \\
wine*$^\dagger$ 	&	\textit{xgboost} 	&	\textbf{0.92}	&	 0.92 	&	 0.9147	&	 0.00 	&	 \textbf{0.94} 	&	 0.7031 	&	 0.08 	&	 0.92 	&	 0.9147 	&	 0.00 \\
\\[-1em]
    \bottomrule
    \end{tabular}
\end{table*}

\subsection{Discussion}

The different error contributions are collectively defined by the fidelity error and approximated through the empirical fidelity error. However, the condition that empirical fidelity error be small is necessary, but not sufficient. Having significant errors in certain regions and none in others may lead to a low error, while altogether not ensuring a good generalization performance. The opposite is also true: a large empirical fidelity error may not lead to a low copy accuracy. Take, for example, errors distributed around the boundary. This may happen when trying to copy a smooth function using linear decision cuts. If errors are very substantial, this may be seen as a problem. However, if the training data are distributed far away from the boundary, errors in this region would have no real impact. No effective error would therefore be measured when substituting the model with the copy.

To a large extent, copy evaluation depends on the available information. The more information we have, the more reliable our estimates will be. If the training data were accessible, we could obtain a direct estimate of the copy generalization performance. Furthermore, we could choose $P_Z$ to be as close to $P$ as possible, \textit{i.e.} redefine the copy operation space to match $P$. If the form of the model was also known, we could refine the choice of copy hypothesis. In those cases where model and copy have similar decision boundary shapes, copying is conducted with greater ease. That is, when the decision function is formed of cuts perpendicular to the axes, \textit{i.e.} it is a random forest, it is easier to copy with a decision tree than it is with a radial basis kernel SVM. Conversely, those models with smooth decision functions are better copied using classifiers other than trees.

At this stage, we may ask ourselves the question: {\it if the training data are available why copy instead of learning a new classifier?} There exist scenarios where a new training may not be advisable. A new model may display very different behaviour and decision properties. This is unacceptable in production environments where performance has to be preserved and controlled. Moreover, training a new classifier with the training data involves having to take care of the overfitting effect. As shown in Sec.~\ref{sec:insights}, when copying we can avoid the hyper-parameter optimization step. 

Another reason to use copies is that when training a new model, we might not be able to recover the same operation point as before. In contrast, as explained in Sec.~\ref{sec:discussion}, a copy can help bias the parameter optimization process towards a desired solution.

In general, copies can be understood as a tool to bridge the gap between accuracy and any other desired property. Copying helps in breaking the trade-offs we face in training high-performance models when characteristics such as interpretability, simplicity or compliance are required.

\section{Applications and limitations}
\label{sec:discussion}

Having demonstrated the feasibility of copying and discussed its main characteristics, in this section we elaborate on its utility in a wide variety of scenarios. We present three use cases with real-life applications of copying. Further, we analyse shortcomings and discuss different approaches to overcoming the identified barriers.  

\subsection{Applications}

One of the main benefits of copying is that it enables differential replication of models. This means that copies can be used to enhance existing solutions. They can, for example, be used to evolve from batch to online learning schemes \cite{Bottou2004}. This extends a model's lifespan as it enables adaptation to data drifts or performance deviations. Equivalently, when new class labels appear during a model's deployment in the wild, copies can account for the new data points and evolve from binary to multiclass classification settings \cite{Oriol2009}. More generally, there are numerous examples were differential replication can be applied to solve specific problems. In the following lines, we describe some of them and discuss how copies could be useful in addressing these issues. \\

\noindent
\textbf{Interpretability.} Recent advances in the field of machine learning have led to increasingly sophisticated models, capable of learning ever more complex problems to a high degree of accuracy. This comes at the cost of simplicity \cite{Doshi-Velez2017TowardsLearning}, \cite{Lipton2016TheInterpretability}, a situation that stands in contrast to the growing demand for transparency in automated processing \cite{EuropeanParliament2016REGULATIONRegulation},\cite{Goodman2017EuropeanExplanation}, \cite{Selbst2017MeaningfulExplanation}. Recent papers have shown that the knowledge acquired by black-box solutions can be transferred to interpretable models such as trees \cite{Bastani2018Inter},\cite{Che2016InterpretablePrediction},\cite{FrosstDistillingTree}, rules \cite{Ribeiro2018Anchors:Explanations} and decision sets \cite{Lakkaraju2016InterpretableSets}. In the copying scenario models of any arbitrary type can be substituted by copies specifically designed to be globally self-explanatory.\\

\begin{figure*}[!ht]
\centering
\subfloat[]{\includegraphics[width=1.67in]{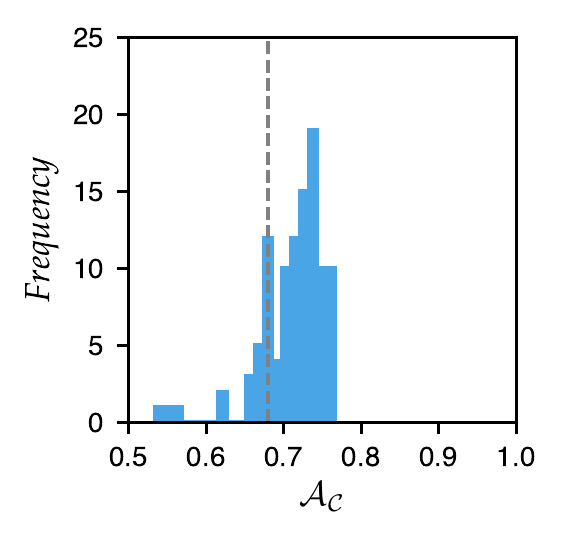}\label{mortgage2.true_copy_acc}}
\subfloat[]{\includegraphics[width=1.5in, trim={0 0.05cm 0 0}]{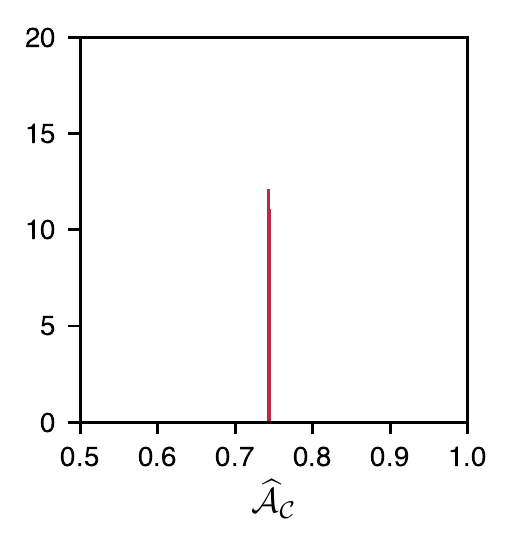}\label{mortgages2.est_copy_acc}}
\subfloat[]{\includegraphics[width=1.5in]{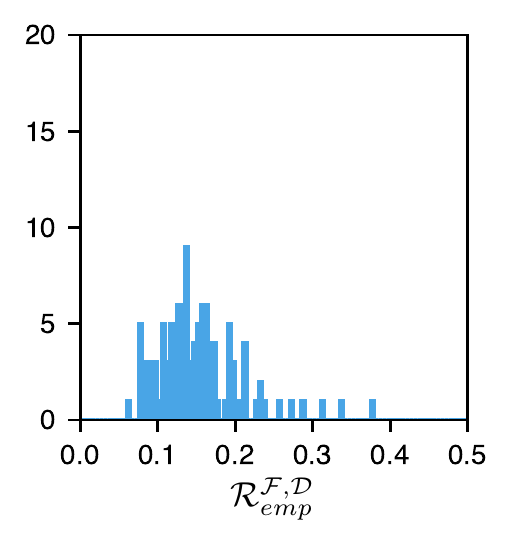}\label{mortgages2.acc_loss}}
\caption{Distribution of values computed for the \textit{scenario 1} for (a) the true copy accuracy, (b) the estimated copy accuracy and (c) the empirical fidelity error over the training data. For comparison purposes, the accuracy of a decision tree trained on original data is shown in black in (a).}
\label{fig:scenario1}
\end{figure*}

\begin{figure*}[!ht]
\centering
\subfloat[]{\includegraphics[width=1.67in]{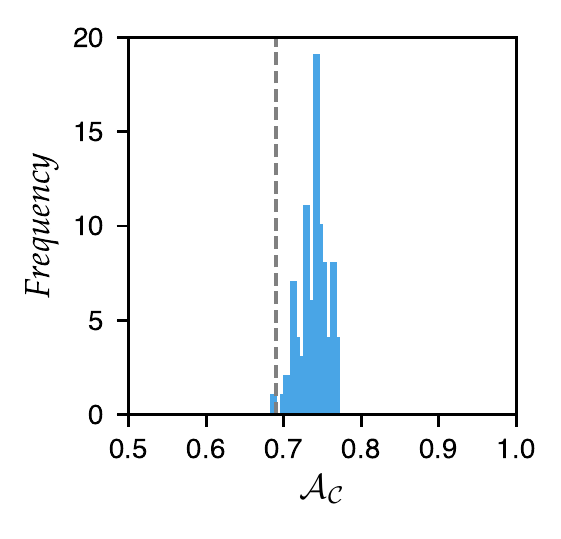}\label{mortgages.true_copy_acc}}
\subfloat[]{\includegraphics[width=1.5in, trim={0 0.05cm 0 0}]{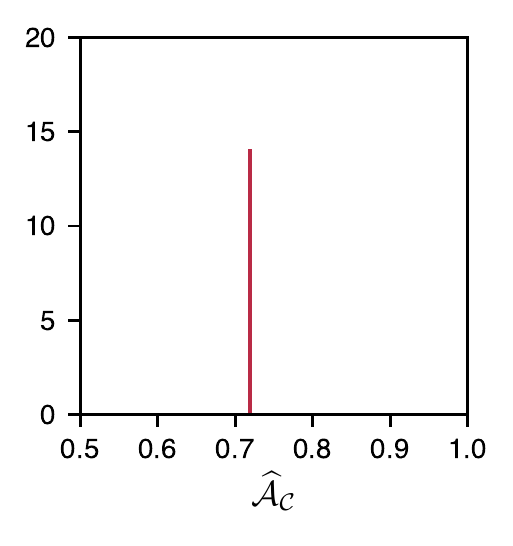}\label{mortgages.est_copy_acc}}
\subfloat[]{\includegraphics[width=1.5in]{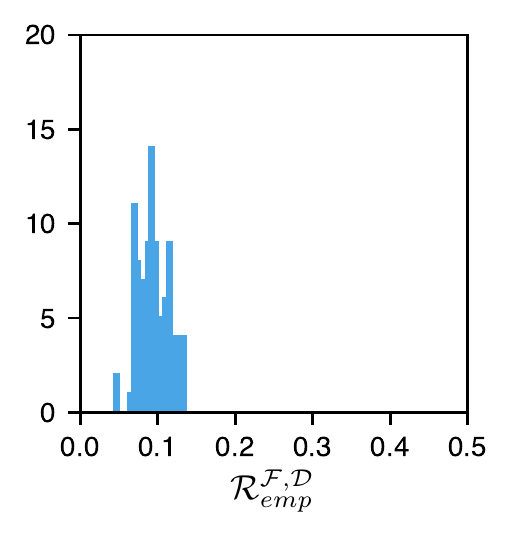}\label{mortgages.acc_loss}}
\caption{Distribution of values computed for the \textit{scenario 2} for (a) the true copy accuracy (b) the estimated copy accuracy and (c) the empirical fidelity error over the training data. For comparison purposes, the accuracy of a decision tree trained on original data is shown in black in (a).}
\label{fig:scenario2}
\end{figure*}

\noindent
\textbf{Production.} Model deployment is often costly in company environments \cite{Sculley2015HiddenSystems}, \cite{Flaounas2017BeyondCompany},  \cite{Spector2012GooglesResearch}, \cite{Zheng2014TheMasses}. Common issues include the inability to maintain the technological infrastructure up-to-date with latest software releases, conflicting versions or incompatible research and deployment environments. Consider the case of neural network library Tensorflow. Despite the library itself provides detailed instructions on how to serve models in production \cite{DeployTensorFlow}, this typically requires several third-party components for docker orchestration, such as Kubernetes or Elastic Container Service \cite{Yau2017HowProduction}, which are seldom compatible with on-premise software infrastructure. Moving to a copy in a less demanding environment helps bridge the gap between the data science and engineering departments. \\

\noindent
\textbf{Fairness and auditing.} Machine learning models can reproduce existing patterns of discrimination \cite{Barocas2016BigImpact}, \cite{Hardt2014HowUnfair}. Some algorithms have been reported to be biased against people with protected characteristics like race \cite{Angwin2016MachineBlacks, Buolamwini2018Gender, Klare2012FaceInformation, Popejoy2016GenomicsDiversity}, gender \cite{Bolukbasi2016ManEmbeddings,  Caliskan2017SemanticsBiases.} or sexual orientation \cite{Guha2010ChallengesSystems}. Under these circumstances distillation has been shown to be useful for model auditing \cite{Tan2018Distill} and so have copies. Upon them, \textit{desiderata} such as equity of learning can be directly imposed to, for example, reduce the biased of trained classifiers.

\subsection{Use cases}

In what follows we demonstrate some of these non-trivial applications in real-life scenarios. First, we derive regulatory-compliant high-performing copies for non-client mortgage loan default prediction in a private dataset from BBVA. Second, we use copies to recover the operation point of a model trained on borrower information from the Lending Club website \cite{LendingKaggle}. Lastly, we study how copies can be applied to obtain a fair classification of alignment in the superheroes dataset~\cite{SuperKaggle}.

\subsubsection{Risk scoring for non-client mortgage loans}

Logistic regression is a widely established technique for credit risk scoring. Mainly because it performs relatively well on credit prediction settings. But also because it offers the additional advantage of a relative ease of interpretation to comply with regulatory requirements. Even so, models based on logistic regression fail to account for non-linearities in the data, which are usually modelled during an increasingly complex preprocessing step. 

During this step, which is critical to maximize business objectives, domain knowledge is exploited to artificially generate a set of highly predictive attributes. Here, a qualified risk analyst is required to conduct a tedious process of trial and error to find an optimal set of variables. This incurs in a large economical cost and a delayed time-to-market delivery. Even worse, preprocessing largely reduces interpretability: new variables often reflect complex relations among attributes and therefore remain non-decomposable \cite{Lipton2016TheInterpretability} as far as the regulators are concerned. 

In what follows, we tackle these issues in two different scenarios. In the first, we use a set of hand-crafted attributes to predict credit default using a logistic regression. We then build a copy that remains interpretable while retaining predictive performance. In the second, we decrease time-to-market delivery by training a high capacity model that avoids the preprocessing step. We copy this model with a simpler architecture that is nonetheless compliant with production and regulatory requirements.

In both cases, we use a private dataset of non-client\footnote[7]{The term non-client refers to those individuals who had no previous contractual relation with the bank at the time of loan application.} mortgage loan applications recorded during 2015 all over Mexico \cite{UncetaNIPS}. This dataset consists of 19 attributes for $1.328$ loan applicants, among which only $77\%$ paid it off.\\

\noindent
\textbf{Deobfuscated risk scoring models.} We emulate a standard production pipeline and preprocess the data to obtain 6 carefully crafted variables. We then train a logistic regression that achieves an accuracy of $0.77$. We copy this whole predictive system, composed of both the preprocessing module and the logistic model, using a decision tree classifier. Fig.~\ref{fig:scenario1} shows the distribution of scores for this experiment. We obtain an averaged copy accuracy of $0.71 \pm 0.04$ and an estimated copy accuracy of $0.74314 \pm 0.00018$. The mean empirical fidelity errors over $\mathscr{Z}$ and $\mathscr{D}$ are $0.03488 \pm 0.00018$ and $0.15 \pm 0.05$, respectively. 

The empirical fidelity error over the synthetic data is small. However, when computed over the original test set this error grows. We argue that if we were to increase the number of synthetic samples, and better explore the boundaries, the approximation error would converge to a more reliable value and the overall error would be reduced.

In this example, the copy uses the deobfuscated 19 variables. Thus, the problem of non-decomposability is effectively solved. For validation purposes, in Fig.~\ref{fig:scenario1}(a) we show the accuracy of a decision tree classifier trained directly on the training data. Note that it is smaller than that of our copy. This shows an additional advantage of copying: it can be used to guide a certain model to a more optimal solution in its parameter space.\\

\noindent
\textbf{High-performance regulatory compliant copies.} In this scenario, we use a high capacity model without any preprocessing. We train a gradient-boosted tree with all the 19 attributes in the training dataset. This model achieves an original accuracy of $0.79$. We copy it using a decision tree classifier and report the results in Fig.~\ref{fig:scenario2}. The mean copy accuracy averaged over all runs is $0.74 \pm 0.02$ and the accuracy estimated using (\ref{eq:estimated}) is equal to $0.7194 \pm 0.0003$. Thus, the average empirical fidelity error is $0.09 \pm 0.0003$ and the average empirical fidelity error over $\mathscr{D}$ is $0.09 \pm 0.02$. Note that while final model attributes differ from this application to that of \textit{scenario\_1}, the same samples are shared in both cases, so as to minimize any bias regarding the specific choice of data.

The difference in performance between the preprocessed logistic model in \textit{scenario\_1} and the copy decision trees in \textit{scenario\_2} is minor when tested against the test data. In Fig.~\ref{fig:scenario2}(a) we display the accuracy achieved by a decision tree trained directly on the training data. This value is equal to $0.69 \pm 0.01$. Comparison between this result and the mean true copy accuracy for this problem provides further evidence for the benefits of using copies in this context. 

\subsubsection{Restoring full operational potential in online loan default prediction}

For predicting whether a potential borrower will repay a loan, the Lending Club website publishes statistics about individual loan applicants \cite{LendingKaggle}. We use these data to show how copies can be used to move a trained classifier to an online setting and recover the original operation point.

\begin{figure*}[!ht]
\centering
\subfloat[]{\includegraphics[width=2.5in]{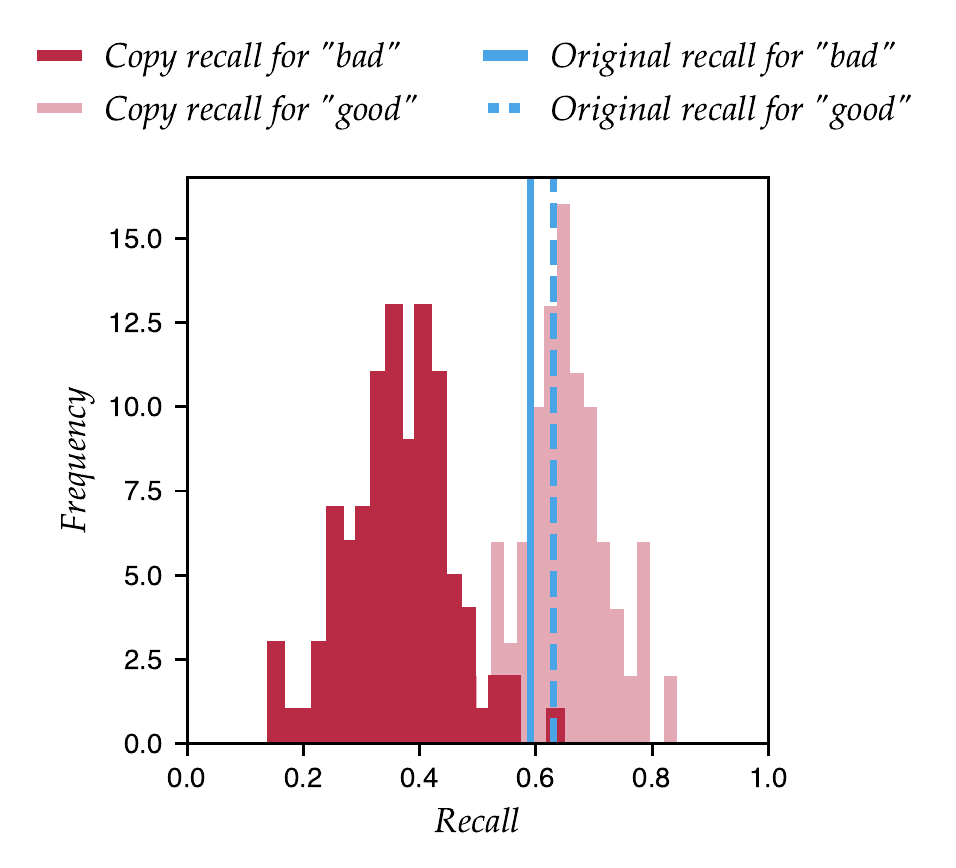}\label{lending_}}
\hspace{1.05cm}
\subfloat[]{\includegraphics[width=2.5in]{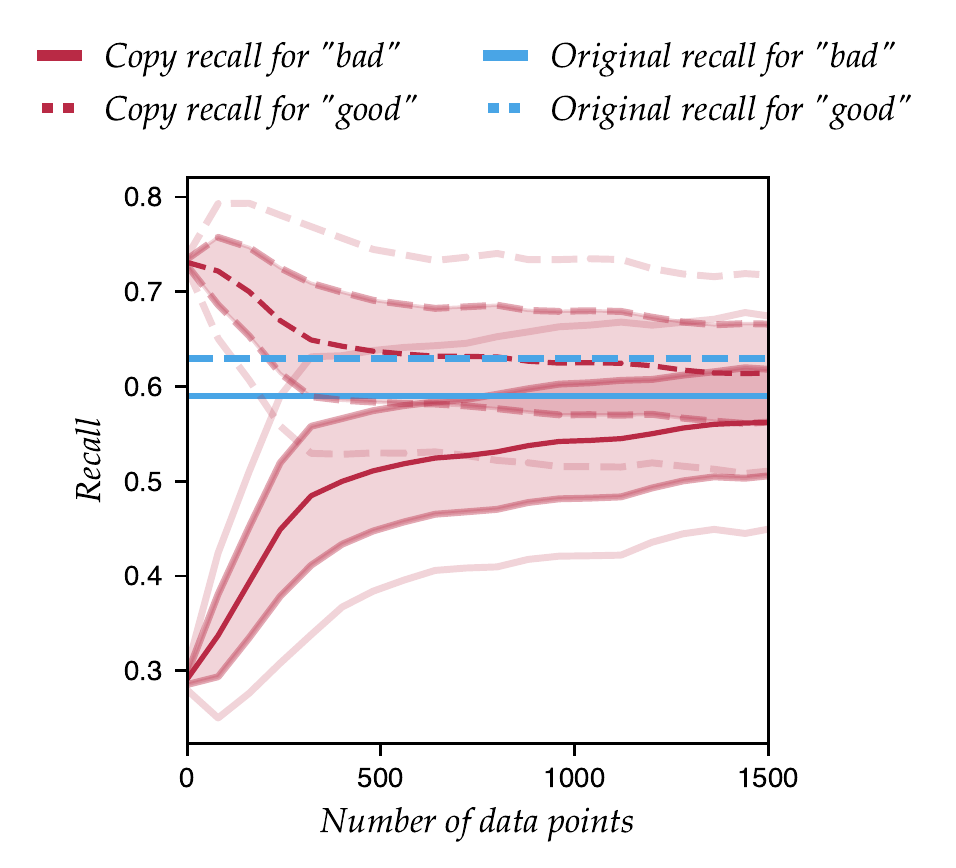}\label{lending_recover}}
\caption{Distribution of (a) recall scores for model and copy and (b) recall scores over the number of new training data points for both classes.}\label{lending}
\end{figure*}

The complete dataset contains a comprehensive list of attributes for all loans issued through the 2007-2015 period, including loan status, latest payment information, number of finance inquires, borrower's annual income or zip code, among others. We remove null and missing values and drop all fields which provide no useful information for inference. We also identify and drop all variables that cause data leakage as those that are typically not available at the time of prediction \cite{Namvar2018CreditEnvironment}. Finally, we label instances by classifying all loans identified as defaulted, charged off or late as \textit{bad}. The resulting database consists of 50 attributes for 887,379 loans, divided into two classes. 

We train a denseNet neural network \cite{Huang2017} consisting of 5 hidden layers with 256, 128, 64, 32 and 16 neurons. We use self-normalizing units\cite{Klambauer2017} to avoid internal covariate shift, a dropout rate of $10\%$ and a least squared loss optimized using Adam. Because  training data are highly imbalanced, with \textit{bad} loans accounting only for $8\%$ of the data, we use balanced batches. We choose our operation point to be that for which the recall values for both classes are closer to each other. Accuracy is equal to $0.63$ and recall is $0.59$ and $0.63$ for the \textit{bad} and \textit{good} classes, respectively.

We copy this model using a neural net with a much simpler architecture, consisting of five fully connected layers with 256, 128, 64, 32 and 16 selu neurons, no dropout and a least-square loss with a default parameter Adam optimizer. We obtain a mean copy accuracy of $0.63 \pm 0.07$. The estimated copy accuracy is $0.603 \pm 0.009$, the empirical fidelity error is $0.042 \pm 0.009$ and the empirical fidelity error over the training data is $0.45 \pm 0.07$. The copy recall distribution over these data is shown in Fig.~\ref{lending}(a), for both classes. We correctly recover the recall operating point for one of the classes, but suffer a loss of around $20\%$ for the other. 

We conclude that we can build copies with online capabilities, while retaining most of the accuracy and reaching a reasonably close operating point. Moreover, in the presence of new data points, copies can be fine tuned to achieve a new desirable operating point, as shown in Fig.~\ref{lending}(b). Here, we recover an equal rate of $59\%$ after visiting a few hundred examples of the training data. It is worth noting that this example also shows that copies can serve as analysis tools for other models. In particular, we observe that the denseNet and the fully connected architectures both have very similar operation points. 

\subsubsection{A fair classification of superhero alignment}

In this use case we exploit a fictitious example that nonetheless represents a use case common to many real scenarios. We assume a model has been trained using protected data attributes and that it cannot be modified to correct for any bias. Instead, we build a copy that reproduces the learned decision function, while excluding these attributes. 

We use superheroes dataset \cite{SuperKaggle}, which describes characteristics such as powers and physical attributes of 660 superheroes in SuperHeroDb \cite{SuperheroDatabase}. We choose alignment as the target attribute to label all superheroes as either \textit{good} or \textit{bad}. We use these data to train a fully-connected artificial neural network with 4 hidden layers, each consisting of 128, 64, 32 and 16 neurons with \textit{SeLu} activation, a softmax cross entropy loss optimized using \textit{Adam} optimizer and a a drop-out equal to $0.6$. This model yields an accuracy of $0.65$

Among the 177 input attributes, \textit{gender} and \textit{race} may be deemed sensitive. The differences in accuracy by the \textit{gender} and \textit{race} groups are shown in Table \ref{tab:acc_group_gender}. In both cases, the resulting decision boundary leads to biased predictions. To overcome this issue, we propose to build a copy that does not include this information.

As a first step, we check that no other variable is correlated with  \textit{gender} and \textit{race} and can leak this information into the copy. We train different models to predict \textit{gender} or \textit{race} using the rest of the variables. We average over 100 runs and obtain a mean balanced accuracy over classes of $0.42 \pm 0.08$ when predicting \textit{gender} and of $0.28 \pm 0.03$ when predicting \textit{race}. We also compute the one-to-one correlation for all attributes. At most, this correlation is equal to $0.18$ in the case of \textit{gender} and to $0.35$ in the case of \textit{race}. We conclude that the remaining attributes are very weakly correlated with these two, so that we can safely remove them without incurring in any leakage of information. 

\begin{table}[ht]
    \centering
    \caption{Accuracy by \textit{gender} and \textit{race} groups for model and copy.}
    \label{tab:acc_group_gender}
    \begin{tabular}{@{}llcc}
    \toprule
    \\[-1em]
        \multirow{1}{2cm}{\footnotesize{Attribute}} &
        \multirow{1}{*}{\footnotesize{Value}} & \multirow{1}{1.5cm}{\centering\footnotesize{Model}} & \multirow{1}{1.5cm}{\centering\footnotesize{Copy}} \\
    \\[-0.75em]
    \toprule
    \\[-0.75em]
        \multirow{2}{*}{\textit{gender}} & \textit{female} & 0.73 & 0.69\\
        & \textit{male} & 0.64 & 0.66 \\
    \midrule
        \multirow{5}{*}{\textit{race}} &
        \textit{human} & 0.78 & 0.76\\
        & \textit{mutant} & 0.75 & 0.75 \\
        & \textit{robot} & 0.67 & 0.5 \\
        & \textit{extraterrestial} & 0.25 & 0.5 \\
        & \textit{other} & 0.59 & 0.64 \\
    \bottomrule
    \end{tabular}
\end{table}

Hence, we extract these two attributes from the synthetic set and build a copy based on the existing network architecture. The mean copy accuracy is $0.66 \pm 0.01$, the estimated copy accuracy is $0.61 \pm 0.02$, and the empirical fidelity error is $0.059 \pm 0.003$. The mean empirical fidelity error over the test data is $0.22 \pm 0.01$. While this value may seem high, we stress that the removal of two variables results in a certain shift of the decision function. As shown in Table \ref{tab:acc_group_gender}, this shift accommodates those instances that are unfairly classified by the model and reduces the overall bias in the copy.  

\subsection{Limitations}

Despite its flexibility and large range of applications, copying has several limitations, for example, when it comes to dealing with high-dimensional data, or with certain problem environments. We highlight some of them.

Copying is highly dependent on the synthetic data generation process. The complexity of this process grows with increasing dimensionality. Hence, while the copying methodology itself remains valid in this context, its performance may be affected. Mostly because sampling an unknown decision function is hard. More so, because we have no information about the training data distribution and lack any insight on how the different classes may be distributed throughout the space. In theory, we could overcome this problem by generating infinite query points. Yet, this is not tractable in practice, since we are limited by our computational resources. 

In our experience, when considering large dimensionality data it is worth replacing uniform sampling distributions with normal distributions. The first conduct an arbitrary exploration of the space, whereas the second better characterize the typicality\footnote{ The concept of typicality refers to properties holding for the vast majority of cases \cite{Volchan}} of a standardized dataset. This is because, as the number of dimensions increases, so do the regions of the space where there are no data present. By using a normal distribution to guide sampling we focus only on those areas that could potentially contain data.

Not only the amount of data but also their structure can be problematic. In structured environments, such as those of images or text, data tend to lie on top of a variety. Finding the optimal synthetic dataset therefore requires sampling the appropriate manifold. While this may be doable, it is not straightforward. In general, copying in such domains would require access to the training data to generate synthetic data with a suitable representation. This could be done, for example, using an autoencoder that ensures image invariance.

An additional limitation is choosing $P_Z$. As shown above, blindly exploring the input space works well for simple cases. As the complexity of the problem grows, however, so does the intricacy of the decision function and more \textit{ad hoc} techniques are needed to appropriately sample the input space. See for example \cite{diego}, where we assay uncertainty based methods to guide sampling,

Lastly, many local minima exist. This is because an infinite number of different synthetic sets can be used to replicate a given decision boundary. In theory, the empirical error is known and equal to zero, so that all sets should converge to the same result. Due to training variability, however, this is not always the case.

\section{Conclusions and future work}
\label{sec:conclusions}

In this paper we propose and validate a model-agnostic framework to copy machine learning classifiers. Copying refers to the process of creating an exact replica of a classifier's decision boundary (or the most similar one if this can not be achieved). As such, this process can be understood as a projection operator of a decision function onto a target model space. The resulting copy optimizes the fidelity measure to preserve the original predictive performance.

We derive the theory for copying and highlight its differences with learning, as traditionally understood by the machine learning community. The process of building a copy does not require access to training data. Moreover, we consider the most general case, where the original model is treated as a black-box whose internals remain unknown. 

We introduce the concept of differential replication as the property of endowing copies with new features by adequately selecting the target projection space. This enables copies to provide reliable solutions to many open issues in machine learning. We also discuss the implications of building copies in practice and introduce a set of performance metrics assuming access to different levels of information. Our experiments demonstrate that our approach is feasible. Moreover, the case studies presented show the potential of copies to ensure interpretability, fairness or productivization of machine learning models.

The problem of representing the decision behaviour of a machine learning model using a finite number of samples is far from being solved. Notably, an in-depth study should be conducted to evaluate methods to sample closed domains where class distribution is governed by an unknown decision function. Much research also remains to be done on how to solve the dual optimization problem. While the single pass-copy provides a reasonable approximation, more general approaches should be studied. 

In this article we restrict ourselves to exploring the application of copies to specific areas such as interpretability, fairness and general enhancement. Nonetheless, there exist other fields were copies are potentially useful. Particularly that of privacy, where copies could be specifically built to be privacy-preserving with respect to the training data. This wide range of applications is ensured by the differential replication property of copies, which enables adaptation to new needs and requirements. This characteristic should be the subject of further research.

\section*{Acknowledgements}

This work has been partially funded by the Spanish project TIN2016-74946-P (MINECO/FEDER, UE), and by AGAUR of the Generalitat de Catalunya through the Industrial PhD grant 2017-DI-25. We gratefully acknowledge the support of BBVA Data \& Analytics for sponsoring the Industrial PhD.

\bibliographystyle{unsrt} 

\appendix
\section{Results for UCI classification}

\begin{table*}[h!]
\centering
\caption{Experimental Results for the 60 Datasets in the UCI Experiment. Blank spaces correspond to cases where models learn a single class label.} \label{tab:UCI_all}
\begin{sideways}
\tiny
    \begin{tabular}{@{}lcccccccccccccc} 
    \toprule
    \\[-1em]
    \multirow{3}{1.5cm}{\scriptsize{Dataset}} &
    \multirow{3}{0.6cm}{\centering\scriptsize{Classes}} & \multirow{3}{0.8cm}{\centering\scriptsize{Samples}} & \multirow{3}{0.8cm}{\centering\scriptsize{Features}} &
    \multirow{3}{0.8cm}{\centering\scriptsize{Original}} & 
    \multirow{3}{0.5cm}{\centering\scriptsize{$\mathcal{A}_{\mathcal{O}}$}}&
    \multicolumn{3}{c}{\multirow{1}{*}{\centering\scriptsize{\textit{decision\_tree}}}} & \multicolumn{3}{c}{\multirow{1}{*}{\centering\scriptsize{\textit{logistic\_regression}}}} &
    \multicolumn{3}{c}{\multirow{1}{*}{\centering\scriptsize{\textit{random\_forest}}}}\\
    \\[-0.5em]
    \cline{7-15}
    \\[-0.75em]
    & & & & & & \multirow{1}{0.4cm}{\centering\scriptsize{$\mathcal{A}_\mathcal{C}$}} & \multirow{1}{0.7cm}{\centering\scriptsize{$\widehat{\mathcal{A}}_\mathcal{C}$}} & \multirow{1}{0.4cm}{\centering\scriptsize{$R_\mathcal{F}^{\mathscr{D}}$}} & \multirow{1}{0.4cm}{\centering\scriptsize{$\mathcal{A}_\mathcal{C}$}} & \multirow{1}{0.7cm}{\centering\scriptsize{$\widehat{\mathcal{A}}_\mathcal{C}$}} & \multirow{1}{0.4cm}{\centering\scriptsize{$R_\mathcal{F}^{\mathscr{D}}$}} &
    \multirow{1}{0.4cm}{\centering\scriptsize{$\mathcal{A}_\mathcal{C}$}} & \multirow{1}{0.7cm}{\centering\scriptsize{$\widehat{\mathcal{A}}_\mathcal{C}$}} & \multirow{1}{0.4cm}{\centering\scriptsize{$R_\mathcal{F}^{\mathscr{D}}$}} \\
    \\[-0.5em]
    \toprule
    \\[-0.75em]
abalone 	&	3	&	3341	&	8	&	 \textit{adaboost} 	&	0.57	&	 0.52 $\pm$ 0.02 	&	 0.5653 $\pm$ 0.0002 	&	 0.27 $\pm$ 0.02 	&	 0.57 $\pm$ 0.00 	&	 0.3266 $\pm$ 0.0001 	&	 0.41 $\pm$ 0.00 	&	 0.58 $\pm$ 0.01 	&	 0.5590 $\pm$ 0.0109 	&	 0.27 $\pm$ 0.01\\
acute-inflammation 	&	2	&	96	&	6	&	 \textit{adaboost} 	&	1	&	 1.00 $\pm$ 0.00 	&	 1.0000 $\pm$ 0.0000 	&	 0.00 $\pm$ 0.00 	&	 1.00 $\pm$ 0.00 	&	 1.0000 $\pm$ 0.0000 	&	 0.00 $\pm$ 0.00 	&	 1.00 $\pm$ 0.00 	&	 1.0000 $\pm$ 0.0000 	&	 0.00 $\pm$ 0.00\\
acute-nephritis 	&	2	&	96	&	6	&	 \textit{adaboost} 	&	1	&	 1.00 $\pm$ 0.00 	&	 1.0000 $\pm$ 0.0000 	&	 0.00 $\pm$ 0.00 	&	 1.00 $\pm$ 0.00 	&	 0.9987 $\pm$ 0.0000 	&	 0.00 $\pm$ 0.00 	&	 1.00 $\pm$ 0.00 	&	 1.0000 $\pm$ 0.0000 	&	 0.00 $\pm$ 0.00\\
bank 	&	2	&	3616	&	16	&	 \textit{adaboost} 	&	0.85	&	 0.82 $\pm$ 0.03 	&	 0.8428 $\pm$ 0.0004 	&	 0.10 $\pm$ 0.03 	&	 0.87 $\pm$ 0.00 	&	 0.5558 $\pm$ 0.0001 	&	 0.12 $\pm$ 0.00 	&	 0.87 $\pm$ 0.00 	&	 0.8353 $\pm$ 0.0026 	&	 0.07 $\pm$ 0.00\\
blood 	&	2	&	598	&	4	&	 \textit{adaboost} 	&	0.71	&	 0.65 $\pm$ 0.04 	&	 0.7000 $\pm$ 0.0001 	&	 0.14 $\pm$ 0.05 	&	 0.67 $\pm$ 0.00 	&	 0.5190 $\pm$ 0.0001 	&	 0.38 $\pm$ 0.00 	&	 0.64 $\pm$ 0.02 	&	 0.7018 $\pm$ 0.0226 	&	 0.16 $\pm$ 0.04\\
breast-cancer 	&	2	&	228	&	9	&	 \textit{adaboost} 	&	0.74	&	 0.74 $\pm$ 0.01 	&	 0.7411 $\pm$ 0.0000 	&	 0.00 $\pm$ 0.01 	&	 0.67 $\pm$ 0.00 	&	 0.4731 $\pm$ 0.0001 	&	 0.31 $\pm$ 0.00 	&	 0.74 $\pm$ 0.00 	&	 0.7412 $\pm$ 0.0000 	&	 0.00 $\pm$ 0.00\\
breast-cancer-wisc 	&	2	&	559	&	9	&	 \textit{adaboost} 	&	\textbf{0.93}	&	 0.93 $\pm$ 0.00 	&	 0.9286 $\pm$ 0.0000 	&	 0.00 $\pm$ 0.00 	&	 \textbf{0.93 $\pm$ 0.00} 	&	 0.6333 $\pm$ 0.0001 	&	 0.06 $\pm$ 0.00 	&	 0.93 $\pm$ 0.00 	&	 0.9286 $\pm$ 0.0000 	&	 0.00 $\pm$ 0.00\\
breast-cancer-wisc-diag 	&	2	&	455	&	30	&	 \textit{adaboost} 	&	0.95	&	 0.95 $\pm$ 0.00 	&	 0.9473 $\pm$ 0.0000 	&	 0.00 $\pm$ 0.00 	&	 0.95 $\pm$ 0.00 	&	 0.7444 $\pm$ 0.0000 	&	 0.04 $\pm$ 0.00 	&	 0.95 $\pm$ 0.00 	&	 0.9473 $\pm$ 0.0000 	&	 0.00 $\pm$ 0.00\\
breast-cancer-wisc-prog 	&	2	&	158	&	33	&	 \textit{adaboost} 	&	0.73	&	 0.72 $\pm$ 0.00 	&	 0.7250 $\pm$ 0.0000 	&	 0.00 $\pm$ 0.00 	&	 0.62 $\pm$ 0.00 	&	 0.5468 $\pm$ 0.0001 	&	 0.35 $\pm$ 0.00 	&	 0.72 $\pm$ 0.00 	&	 0.7249 $\pm$ 0.0000 	&	 0.00 $\pm$ 0.00\\
breast-tissue 	&	6	&	84	&	9	&	 \textit{adaboost} 	&	0.59	&	 0.61 $\pm$ 0.02 	&	 0.5909 $\pm$ 0.0000 	&	 0.16 $\pm$ 0.04 	&	 0.36 $\pm$ 0.00 	&	 0.3838 $\pm$ 0.0001 	&	 0.59 $\pm$ 0.00 	&	 0.57 $\pm$ 0.02 	&	 0.5909 $\pm$ 0.0223 	&	 0.18 $\pm$ 0.01\\
chess-krvkp 	&	2	&	2556	&	36	&	 \textit{ann} 	&	0.99	&	 0.89 $\pm$ 0.02 	&	 0.9527 $\pm$ 0.0002 	&	 0.11 $\pm$ 0.02 	&	 0.91 $\pm$ 0.00 	&	 0.9603 $\pm$ 0.0000 	&	 0.10 $\pm$ 0.00 	&	 0.91 $\pm$ 0.01 	&	 0.9670 $\pm$ 0.0091 	&	 0.09 $\pm$ 0.01\\
congressional-voting 	&	2	&	348	&	16	&	 \textit{ann} 	&	0.61	&	 0.61 $\pm$ 0.00 	&	 0.6092 $\pm$ 0.0000 	&	 0.00 $\pm$ 0.00 	&	 0.61 $\pm$ 0.00 	&	 0.6087 $\pm$ 0.0000 	&	 0.00 $\pm$ 0.00 	&	 0.61 $\pm$ 0.00 	&	 0.6092 $\pm$ 0.0000 	&	 0.00 $\pm$ 0.00\\
conn-bench-sonar-mines-rocks 	&	2	&	166	&	60	&	 \textit{ann} 	&	0.88	&	 0.58 $\pm$ 0.08 	&	 0.5897 $\pm$ 0.0004 	&	 0.40 $\pm$ 0.07 	&	 0.81 $\pm$ 0.01 	&	 0.8079 $\pm$ 0.0000 	&	 0.12 $\pm$ 0.01 	&	 0.69 $\pm$ 0.03 	&	 0.7405 $\pm$ 0.0327 	&	 0.21 $\pm$ 0.04\\
connect-4 	&	2	&	54045	&	42	&	 \textit{ann} 	&	0.87	&	 0.54 $\pm$ 0.04 	&	 0.6518 $\pm$ 0.0003 	&	 0.46 $\pm$ 0.04 	&	 0.60 $\pm$ 0.00 	&	 0.6871 $\pm$ 0.0000 	&	 0.39 $\pm$ 0.00 	&	 0.66 $\pm$ 0.00 	&	 0.7442 $\pm$ 0.0031 	&	 0.32 $\pm$ 0.00\\
contrac 	&	3	&	1178	&	9	&	 \textit{ann} 	&	0.55	&	 0.55 $\pm$ 0.01 	&	 0.5251 $\pm$ 0.0001 	&	 0.07 $\pm$ 0.02 	&	 0.50 $\pm$ 0.00 	&	 0.4111 $\pm$ 0.0000 	&	 0.40 $\pm$ 0.00 	&	 0.55 $\pm$ 0.00 	&	 0.5357 $\pm$ 0.0043 	&	 0.03 $\pm$ 0.01\\
credit-approval 	&	2	&	552	&	15	&	 \textit{ann} 	&	0.79	&	 0.80 $\pm$ 0.02 	&	 0.7268 $\pm$ 0.0003 	&	 0.10 $\pm$ 0.02 	&	 0.72 $\pm$ 0.00 	&	 0.7066 $\pm$ 0.0000 	&	 0.19 $\pm$ 0.00 	&	 0.79 $\pm$ 0.01 	&	 0.7527 $\pm$ 0.0071 	&	 0.03 $\pm$ 0.01\\
cylinder-bands 	&	2	&	409	&	35	&	 \textit{ann} 	&	0.69	&	 0.60 $\pm$ 0.05 	&	 0.6083 $\pm$ 0.0002 	&	 0.37 $\pm$ 0.05 	&	 0.71 $\pm$ 0.00 	&	 0.6354 $\pm$ 0.0000 	&	 0.25 $\pm$ 0.00 	&	 0.65 $\pm$ 0.01 	&	 0.6445 $\pm$ 0.0121 	&	 0.35 $\pm$ 0.01\\
echocardiogram 	&	2	&	104	&	10	&	 \textit{ann} 	&	\textbf{0.3}	&	 0.33 $\pm$ 0.04 	&	 0.2879 $\pm$ 0.0000 	&	 0.05 $\pm$ 0.03 	&	 0.30 $\pm$ 0.00 	&	 0.2960 $\pm$ 0.0000 	&	 0.00 $\pm$ 0.00 	&	 0.30 $\pm$ 0.00 	&	 0.2922 $\pm$ 0.0000 	&	 0.00 $\pm$ 0.00\\
energy-y1 	&	3	&	614	&	8	&	 \textit{ann} 	&	\textbf{0.96}	&	 \textbf{0.96 $\pm$ 0.01} 	&	 0.9537 $\pm$ 0.0001 	&	 0.00 $\pm$ 0.01 	&	 \textbf{0.78 $\pm$ 0.00} 	&	 0.7744 $\pm$ 0.0001 	&	 0.23 $\pm$ 0.00 	&	 \textbf{0.96 $\pm$ 0.00} 	&	 0.9551 $\pm$ 0.0039 	&	 0.00 $\pm$ 0.00\\
energy-y2 	&	3	&	614	&	8	&	 \textit{ann} 	&	0.84	&	 0.84 $\pm$ 0.00 	&	 0.8403 $\pm$ 0.0001 	&	 0.00 $\pm$ 0.00 	&	 0.79 $\pm$ 0.00 	&	 0.7835 $\pm$ 0.0001 	&	 0.09 $\pm$ 0.00 	&	 0.84 $\pm$ 0.01 	&	 0.8410 $\pm$ 0.0065 	&	 0.01 $\pm$ 0.00\\
fertility 	&	2	&	80	&	9	&	 \textit{random\_forest} 	&	0.9	&	 0.90 $\pm$ 0.00 	&	 0.8993 $\pm$ 0.0000 	&	 0.00 $\pm$ 0.00 	&	 0.75 $\pm$ 0.00 	&	 0.8255 $\pm$ 0.0000 	&	 0.15 $\pm$ 0.00 	&	 0.90 $\pm$ 0.00 	&	 0.8993 $\pm$ 0.0000 	&	 0.00 $\pm$ 0.00\\
haberman-survival 	&	2	&	244	&	3	&	 \textit{random\_forest} 	&	0.61	&	 0.61 $\pm$ 0.01 	&	 0.6125 $\pm$ 0.0000 	&	 0.01 $\pm$ 0.01 	&	 0.61 $\pm$ 0.00 	&	 0.4888 $\pm$ 0.0001 	&	 0.26 $\pm$ 0.00 	&	 0.61 $\pm$ 0.01 	&	 0.6123 $\pm$ 0.0074 	&	 0.00 $\pm$ 0.01\\
heart-hungarian 	&	2	&	235	&	12	&	 \textit{random\_forest} 	&	0.76	&	 0.79 $\pm$ 0.04 	&	 0.7445 $\pm$ 0.0002 	&	 0.07 $\pm$ 0.03 	&	 0.81 $\pm$ 0.00 	&	 0.6457 $\pm$ 0.0001 	&	 0.05 $\pm$ 0.00 	&	 0.80 $\pm$ 0.00 	&	 0.7498 $\pm$ 0.0000 	&	 0.03 $\pm$ 0.00\\
hepatitis 	&	2	&	124	&	19	&	 \textit{random\_forest} 	&	0.74	&	 0.74 $\pm$ 0.05 	&	 0.6991 $\pm$ 0.0002 	&	 0.05 $\pm$ 0.04 	&	 0.90 $\pm$ 0.00 	&	 0.5973 $\pm$ 0.0000 	&	 0.16 $\pm$ 0.00 	&	 0.75 $\pm$ 0.01 	&	 0.7145 $\pm$ 0.0097 	&	 0.00 $\pm$ 0.01\\
ilpd-indian-liver 	&	2	&	466	&	9	&	 \textit{random\_forest} 	&	0.62	&	 0.59 $\pm$ 0.02 	&	 0.6207 $\pm$ 0.0001 	&	 0.36 $\pm$ 0.02 	&	 0.56 $\pm$ 0.00 	&	 0.6020 $\pm$ 0.0000 	&	 0.47 $\pm$ 0.00 	&	 0.62 $\pm$ 0.01 	&	 0.6224 $\pm$ 0.0089 	&	 0.37 $\pm$ 0.01\\
ionosphere 	&	2	&	280	&	33	&	 \textit{random\_forest} 	&	0.94	&	 0.92 $\pm$ 0.03 	&	 0.8543 $\pm$ 0.0004 	&	 0.07 $\pm$ 0.03 	&	 0.89 $\pm$ 0.00 	&	 0.7793 $\pm$ 0.0000 	&	 0.11 $\pm$ 0.00 	&	 0.95 $\pm$ 0.01 	&	 0.8867 $\pm$ 0.0138 	&	 0.05 $\pm$ 0.01\\
iris 	&	3	&	120	&	4	&	 \textit{random\_forest} 	&	0.93	&	 \textbf{0.95 $\pm$ 0.02} 	&	 0.9332 $\pm$ 0.0000 	&	 0.02 $\pm$ 0.02 	&	 \textbf{0.70 $\pm$ 0.00} 	&	 0.6778 $\pm$ 0.0003 	&	 0.30 $\pm$ 0.00 	&	 \textbf{0.95 $\pm$ 0.02} 	&	 0.9333 $\pm$ 0.0163 	&	 0.02 $\pm$ 0.02\\
magic 	&	2	&	15216	&	10	&	 \textit{random\_forest} 	&	0.88	&	 0.83 $\pm$ 0.01 	&	 0.8024 $\pm$ 0.0003 	&	 0.10 $\pm$ 0.01 	&	 0.79 $\pm$ 0.00 	&	 0.6809 $\pm$ 0.0001 	&	 0.17 $\pm$ 0.00 	&	 0.86 $\pm$ 0.00 	&	 0.8316 $\pm$ 0.0015 	&	 0.06 $\pm$ 0.00\\
mammographic 	&	2	&	768	&	5	&	 \textit{random\_forest} 	&	0.8	&	 0.80 $\pm$ 0.00 	&	 0.7974 $\pm$ 0.0001 	&	 0.01 $\pm$ 0.00 	&	 0.76 $\pm$ 0.00 	&	 0.5377 $\pm$ 0.0001 	&	 0.19 $\pm$ 0.00 	&	 0.80 $\pm$ 0.00 	&	 0.7974 $\pm$ 0.0039 	&	 0.01 $\pm$ 0.00\\
miniboone 	&	2	&	104051	&	50	&	 \textit{random\_forest} 	&	0.94	&	 0.86 $\pm$ 0.01 	&	 0.8402 $\pm$ 0.0004 	&	 0.12 $\pm$ 0.01 	&	 0.84 $\pm$ 0.00 	&	 0.6951 $\pm$ 0.0004 	&	 0.15 $\pm$ 0.00 	&	 0.90 $\pm$ 0.00 	&	 0.8679 $\pm$ 0.0011 	&	 0.07 $\pm$ 0.00\\
molec-biol-splice 	&	3	&	2552	&	60	&	 \textit{linear\_svm} 	&	0.84	&	 0.77 $\pm$ 0.01 	&	 0.7151 $\pm$ 0.0004 	&	 0.17 $\pm$ 0.01 	&	 0.77 $\pm$ 0.00 	&	 0.7801 $\pm$ 0.0001 	&	 0.14 $\pm$ 0.00 	&	 0.80 $\pm$ 0.00 	&	 0.7590 $\pm$ 0.0040 	&	 0.15 $\pm$ 0.00\\
mushroom 	&	2	&	6499	&	21	&	 \textit{linear\_svm} 	&	0.98	&	 0.95 $\pm$ 0.02 	&	 0.9526 $\pm$ 0.0002 	&	 0.03 $\pm$ 0.02 	&	 0.98 $\pm$ 0.00 	&	 0.9777 $\pm$ 0.0000 	&	 0.00 $\pm$ 0.00 	&	 0.87 $\pm$ 0.05 	&	 0.9549 $\pm$ 0.0544 	&	 0.11 $\pm$ 0.06\\
musk-1 	&	2	&	380	&	166	&	 \textit{linear\_svm} 	&	\textbf{0.88}	&	 0.54 $\pm$ 0.04 	&	 0.5620 $\pm$ 0.0003 	&	 0.46 $\pm$ 0.05 	&	 0.88 $\pm$ 0.00 	&	 0.8732 $\pm$ 0.0000 	&	 0.01 $\pm$ 0.00 	&	 \textbf{0.67 $\pm$ 0.06} 	&	 0.7323 $\pm$ 0.0577 	&	 0.32 $\pm$ 0.04\\
musk-2 	&	2	&	5278	&	166	&	 \textit{linear\_svm} 	&	\textbf{0.96}	&	 0.50 $\pm$ 0.05 	&	 0.6005 $\pm$ 0.0004 	&	 0.50 $\pm$ 0.05 	&	 0.96 $\pm$ 0.00 	&	 0.9556 $\pm$ 0.0000 	&	 0.00 $\pm$ 0.00 	&	 \textbf{0.56 $\pm$ 0.04} 	&	 0.7745 $\pm$ 0.0375 	&	 0.44 $\pm$ 0.04\\
oocytes\_merluccius\_nucleus\_4d 	&	2	&	817	&	41	&	 \textit{linear\_svm} 	&	0.82	&	 0.47 $\pm$ 0.06 	&	 0.6460 $\pm$ 0.0003 	&	 0.52 $\pm$ 0.06 	&	 0.81 $\pm$ 0.00 	&	 0.8144 $\pm$ 0.0000 	&	 0.00 $\pm$ 0.00 	&	 0.59 $\pm$ 0.03 	&	 0.7317 $\pm$ 0.0264 	&	 0.38 $\pm$ 0.03\\
oocytes\_trisopterus\_nucleus\_2f 	&	2	&	729	&	25	&	 \textit{linear\_svm} 	&	0.81	&	 0.56 $\pm$ 0.05 	&	 0.6946 $\pm$ 0.0002 	&	 0.43 $\pm$ 0.07 	&	 0.81 $\pm$ 0.00 	&	 0.8084 $\pm$ 0.0000 	&	 0.00 $\pm$ 0.00 	&	 0.63 $\pm$ 0.03 	&	 0.7543 $\pm$ 0.0296 	&	 0.32 $\pm$ 0.03\\
parkinsons 	&	2	&	156	&	22	&	 \textit{linear\_svm} 	&	0.9	&	 0.83 $\pm$ 0.04 	&	 0.7990 $\pm$ 0.0003 	&	 0.11 $\pm$ 0.06 	&	 0.90 $\pm$ 0.00 	&	 0.8950 $\pm$ 0.0000 	&	 0.00 $\pm$ 0.00 	&	 0.89 $\pm$ 0.01 	&	 0.8562 $\pm$ 0.0126 	&	 0.02 $\pm$ 0.02\\
pima 	&	2	&	614	&	8	&	 \textit{linear\_svm} 	&	0.72	&	 0.72 $\pm$ 0.01 	&	 0.6967 $\pm$ 0.0001 	&	 0.04 $\pm$ 0.01 	&	 0.72 $\pm$ 0.00 	&	 0.7203 $\pm$ 0.0000 	&	 0.00 $\pm$ 0.00 	&	 0.71 $\pm$ 0.01 	&	 0.7097 $\pm$ 0.0050 	&	 0.02 $\pm$ 0.01\\
pittsburg-bridges-MATERIAL 	&	3	&	84	&	7	&	 \textit{linear\_svm} 	&	0.91	&	 0.91 $\pm$ 0.00 	&	 0.9090 $\pm$ 0.0000 	&	 0.00 $\pm$ 0.00 	&	 0.91 $\pm$ 0.00 	&	 0.8968 $\pm$ 0.0000 	&	 0.00 $\pm$ 0.00 	&	 0.91 $\pm$ 0.00 	&	 0.9091 $\pm$ 0.0000 	&	 0.00 $\pm$ 0.00\\
pittsburg-bridges-REL-L 	&	3	&	82	&	7	&	 \textit{linear\_svm} 	&	\textbf{0.67}	&	 0.67 $\pm$ 0.00 	&	 0.6667 $\pm$ 0.0000 	&	 0.00 $\pm$ 0.00 	&	 0.67 $\pm$ 0.00 	&	 0.6667 $\pm$ 0.0000 	&	 0.00 $\pm$ 0.00 	&	 0.67 $\pm$ 0.00 	&	 0.6667 $\pm$ 0.0000 	&	 0.00 $\pm$ 0.00\\
pittsburg-bridges-T-OR-D 	&	2	&	81	&	7	&	 \textit{rbf\_svm} 	&	0.86	&	 0.86 $\pm$ 0.00 	&	 0.8562 $\pm$ 0.0000 	&	 0.00 $\pm$ 0.00 	&	 0.90 $\pm$ 0.00 	&	 0.6932 $\pm$ 0.0001 	&	 0.24 $\pm$ 0.00 	&	 0.86 $\pm$ 0.00 	&	 0.8568 $\pm$ 0.0000 	&	 0.00 $\pm$ 0.00\\
planning 	&	2	&	145	&	12	&	 \textit{rbf\_svm} 	&	\textbf{0.7}	&	 0.70 $\pm$ 0.00 	&	 0.7027 $\pm$ 0.0000 	&	 0.00 $\pm$ 0.00 	&	-	&	-	&	-	&	 0.70 $\pm$ 0.00 	&	 0.7027 $\pm$ 0.0000 	&	 0.00 $\pm$ 0.00\\
ringnorm 	&	2	&	5920	&	18	&	 \textit{rbf\_svm} 	&	0.98	&	 0.88 $\pm$ 0.01 	&	 0.7992 $\pm$ 0.0005 	&	 0.11 $\pm$ 0.01 	&	 0.74 $\pm$ 0.00 	&	 0.7049 $\pm$ 0.0000 	&	 0.26 $\pm$ 0.00 	&	 0.95 $\pm$ 0.00 	&	 0.8877 $\pm$ 0.0022 	&	 0.05 $\pm$ 0.00\\
seeds 	&	3	&	168	&	7	&	 \textit{rbf\_svm} 	&	0.88	&	 0.90 $\pm$ 0.02 	&	 0.8006 $\pm$ 0.0003 	&	 0.06 $\pm$ 0.03 	&	 0.88 $\pm$ 0.00 	&	 0.8582 $\pm$ 0.0001 	&	 0.00 $\pm$ 0.00 	&	 0.92 $\pm$ 0.02 	&	 0.8487 $\pm$ 0.0158 	&	 0.04 $\pm$ 0.02\\
spambase 	&	2	&	3680	&	57	&	 \textit{rbf\_svm} 	&	0.93	&	 0.45 $\pm$ 0.10 	&	 0.7610 $\pm$ 0.0002 	&	 0.55 $\pm$ 0.11 	&	 0.92 $\pm$ 0.00 	&	 0.9227 $\pm$ 0.0000 	&	 0.02 $\pm$ 0.00 	&	 0.53 $\pm$ 0.08 	&	 0.8577 $\pm$ 0.0824 	&	 0.48 $\pm$ 0.09\\
statlog-australian-credit 	&	2	&	552	&	14	&	 \textit{rbf\_svm} 	&	\textbf{0.68}	&	 0.68 $\pm$ 0.00 	&	 0.6812 $\pm$ 0.0000 	&	 0.00 $\pm$ 0.00 	&	-	&	-	&	-	&	 0.68 $\pm$ 0.00 	&	 0.6812 $\pm$ 0.0000 	&	 0.00 $\pm$ 0.00\\
statlog-german-credit 	&	2	&	800	&	24	&	 \textit{rbf\_svm} 	&	0.79	&	 0.59 $\pm$ 0.03 	&	 0.6820 $\pm$ 0.0003 	&	 0.36 $\pm$ 0.04 	&	 0.78 $\pm$ 0.00 	&	 0.7817 $\pm$ 0.0000 	&	 0.02 $\pm$ 0.00 	&	 0.76 $\pm$ 0.01 	&	 0.7369 $\pm$ 0.0121 	&	 0.07 $\pm$ 0.01\\
statlog-heart 	&	2	&	216	&	13	&	 \textit{rbf\_svm} 	&	0.85	&	 0.81 $\pm$ 0.02 	&	 0.8051 $\pm$ 0.0002 	&	 0.05 $\pm$ 0.01 	&	 0.85 $\pm$ 0.00 	&	 0.8510 $\pm$ 0.0000 	&	 0.00 $\pm$ 0.00 	&	 0.81 $\pm$ 0.01 	&	 0.8271 $\pm$ 0.0056 	&	 0.04 $\pm$ 0.01\\
statlog-image 	&	7	&	1848	&	18	&	 \textit{rbf\_svm} 	&	0.95	&	 0.74 $\pm$ 0.01 	&	 0.8351 $\pm$ 0.0003 	&	 0.25 $\pm$ 0.02 	&	 0.74 $\pm$ 0.00 	&	 0.8218 $\pm$ 0.0001 	&	 0.25 $\pm$ 0.00 	&	 0.78 $\pm$ 0.01 	&	 0.8993 $\pm$ 0.0076 	&	 0.21 $\pm$ 0.01\\
statlog-vehicle 	&	4	&	676	&	18	&	 \textit{rbf\_svm} 	&	0.85	&	 0.56 $\pm$ 0.05 	&	 0.5489 $\pm$ 0.0002 	&	 0.40 $\pm$ 0.04 	&	 0.66 $\pm$ 0.00 	&	 0.6618 $\pm$ 0.0000 	&	 0.33 $\pm$ 0.00 	&	 0.65 $\pm$ 0.02 	&	 0.6971 $\pm$ 0.0186 	&	 0.29 $\pm$ 0.02\\
synthetic-control 	&	6	&	480	&	60	&	 \textit{xgboost} 	&	0.96	&	 0.75 $\pm$ 0.02 	&	 0.6547 $\pm$ 0.0007 	&	 0.23 $\pm$ 0.03 	&	 0.81 $\pm$ 0.00 	&	 0.6096 $\pm$ 0.0001 	&	 0.22 $\pm$ 0.00 	&	 0.94 $\pm$ 0.01 	&	 0.7683 $\pm$ 0.0095 	&	 0.02 $\pm$ 0.01\\
teaching 	&	3	&	120	&	5	&	 \textit{xgboost} 	&	0.55	&	 0.55 $\pm$ 0.02 	&	 0.5476 $\pm$ 0.0000 	&	 0.01 $\pm$ 0.01 	&	 0.35 $\pm$ 0.00 	&	 0.3047 $\pm$ 0.0002 	&	 0.65 $\pm$ 0.00 	&	 0.55 $\pm$ 0.01 	&	 0.5476 $\pm$ 0.0097 	&	 0.00 $\pm$ 0.01\\
tic-tac-toe 	&	2	&	766	&	9	&	 \textit{xgboost} 	&	0.97	&	 0.97 $\pm$ 0.00 	&	 0.9740 $\pm$ 0.0000 	&	 0.00 $\pm$ 0.00 	&	 0.71 $\pm$ 0.00 	&	 0.8224 $\pm$ 0.0002 	&	 0.26 $\pm$ 0.00 	&	 0.97 $\pm$ 0.00 	&	 0.9740 $\pm$ 0.0000 	&	 0.00 $\pm$ 0.00\\
titanic 	&	2	&	1760	&	3	&	 \textit{xgboost} 	&	0.78	&	 0.78 $\pm$ 0.00 	&	 0.7778 $\pm$ 0.0000 	&	 0.00 $\pm$ 0.00 	&	 0.76 $\pm$ 0.00 	&	 0.5837 $\pm$ 0.0000 	&	 0.10 $\pm$ 0.00 	&	 0.78 $\pm$ 0.00 	&	 0.7778 $\pm$ 0.0000 	&	 0.00 $\pm$ 0.00\\
twonorm 	&	2	&	5920	&	20	&	 \textit{xgboost} 	&	0.98	&	 0.87 $\pm$ 0.01 	&	 0.7454 $\pm$ 0.0003 	&	 0.13 $\pm$ 0.01 	&	 0.98 $\pm$ 0.00 	&	 0.8913 $\pm$ 0.0000 	&	 0.02 $\pm$ 0.00 	&	 0.96 $\pm$ 0.00 	&	 0.8842 $\pm$ 0.0021 	&	 0.02 $\pm$ 0.00\\
vertebral-column-2clases 	&	2	&	248	&	6	&	 \textit{xgboost} 	&	0.77	&	 0.78 $\pm$ 0.02 	&	 0.7669 $\pm$ 0.0001 	&	 0.05 $\pm$ 0.02 	&	 0.81 $\pm$ 0.01 	&	 0.7244 $\pm$ 0.0001 	&	 0.14 $\pm$ 0.01 	&	 0.80 $\pm$ 0.01 	&	 0.7687 $\pm$ 0.0074 	&	 0.02 $\pm$ 0.01\\
vertebral-column-3clases 	&	3	&	248	&	6	&	 \textit{xgboost} 	&	0.84	&	 0.84 $\pm$ 0.02 	&	 0.8379 $\pm$ 0.0000 	&	 0.02 $\pm$ 0.01 	&	 0.80 $\pm$ 0.00 	&	 0.7575 $\pm$ 0.0000 	&	 0.10 $\pm$ 0.00 	&	 0.84 $\pm$ 0.00 	&	 0.8378 $\pm$ 0.0000 	&	 0.00 $\pm$ 0.00\\
waveform 	&	3	&	4000	&	21	&	 \textit{xgboost} 	&	0.84	&	 0.77 $\pm$ 0.01 	&	 0.6250 $\pm$ 0.0005 	&	 0.18 $\pm$ 0.01 	&	 0.85 $\pm$ 0.00 	&	 0.7053 $\pm$ 0.0000 	&	 0.09 $\pm$ 0.00 	&	 0.83 $\pm$ 0.00 	&	 0.7318 $\pm$ 0.0044 	&	 0.08 $\pm$ 0.00\\
waveform-noise 	&	3	&	4000	&	40	&	 \textit{xgboost} 	&	0.84	&	 0.76 $\pm$ 0.01 	&	 0.6610 $\pm$ 0.0004 	&	 0.19 $\pm$ 0.01 	&	 0.87 $\pm$ 0.00 	&	 0.7083 $\pm$ 0.0000 	&	 0.08 $\pm$ 0.00 	&	 0.85 $\pm$ 0.00 	&	 0.7415 $\pm$ 0.0043 	&	 0.07 $\pm$ 0.00\\
wine 	&	3	&	142	&	11	&	 \textit{xgboost} 	&	\textbf{0.92}	&	 0.92 $\pm$ 0.00 	&	 0.9147 $\pm$ 0.0000 	&	 0.00 $\pm$ 0.00 	&	 \textbf{0.94 $\pm$ 0.00} 	&	 0.7031 $\pm$ 0.0001 	&	 0.08 $\pm$ 0.00 	&	 0.92 $\pm$ 0.00 	&	 0.9147 $\pm$ 0.0000 	&	 0.00 $\pm$ 0.00\\
\\[-1em]
    \bottomrule
    \end{tabular}
\end{sideways}
\end{table*}

\end{document}